\documentclass[journal,transmag]{IEEEtran}

\makeatletter
\def\endthebibliography{%
	\def\@noitemerr{\@latex@warning{Empty `thebibliography' environment}}%
	\endlist
}
\makeatother

\usepackage[utf8x]{inputenc}
\usepackage[T1]{fontenc}
\usepackage{supertabular,booktabs}
\usepackage{longtable}
\usepackage{graphicx}

\begin{document}
%
\title{Heuristic and Metaheuristic Methods for the Unrelated Machines Scheduling Problem: A Survey}


\author{\IEEEauthorblockN{Marko Đurasević\IEEEauthorrefmark{1},
Domagoj Jakobović\IEEEauthorrefmark{1}}
\IEEEauthorblockA{\IEEEauthorrefmark{1}University of Zagreb, Faculty of Electrical Engineering and Computing}
\thanks{Corresponding author: M. Đurasević (email: marko.durasevic@fer.hr).}}

\markboth{IEEE Transactions on Cybernetics}%
{Shell \MakeLowercase{\textit{et al.}}: Bare Demo of IEEEtran.cls for IEEE Transactions on Magnetics Journals}

\maketitle

	\begin{abstract}
Today scheduling problems have an immense effect on various areas of human lives, be it from their application in manufacturing and production industry, transportation, or workforce allocation. The unrelated parallel machines scheduling problem (UPMSP), which is only one of the many different problem types that exist, found its application in many areas like production industries or distributed computing. Due to the complexity of the problem, heuristic and metaheuristic methods are gaining more attention for solving it. Although this problem variant did not receive much attention as other models, recent years saw the increase of research dealing with this problem. During that time, many different problem variants, solution methods, or other interesting research directions were considered. However, no study has until now tried to systematise the research in which heuristic methods are applied for the UPMSP. The goal of this study is to provide an extensive literature review on the application of heuristic and metaheuristic methods for solving the UPMSP. The research was systematised and classified into several categories to enable an easy overview of the different problem and solution variants. Additionally, current trends and possible future research directions are also shortly outlined.
	\end{abstract}
	

\section{Introduction}
Scheduling is usually defined as the process of allocating activities to scarce resources to optimise one or more user defined criteria \cite{leung2004}. It takes many forms and appears in a multitude of real world situation. Some examples include scheduling in manufacturing and different industries, processes in cloud and grid environments, workers in different industries (nurses, drivers, airline crews), and more \cite{leung2004, Pinedo2008}. Therefore, it is easy to see that scheduling directly affects the everyday life of most people. As such, it is of great importance to improve the efficiency of scheduling methods not only to increase the production and decrease the cost in manufacturing, but also to improve the general satisfaction of people and quality of life. Therefore, a great deal of research has focused on developing new and improving existing methods to more efficiently solve various scheduling problems. 

Scheduling problems can be solved using different methods, which can be roughly grouped into three categories: exact, approximate, and heuristic. Exact methods provide an optimal solution to the problem by traversing the entire search space and guaranteeing that a better solution does not exist. However, such methods are computationally expensive, which makes them useful only for smaller problems and in situations where the time to obtain solutions is not critical. On the other hand, approximation methods have a smaller computational complexity than exact methods, but provide a guarantee that the solution obtained by them is within a given margin from the optimal solution. These methods are difficult to design as they need to be mathematically well defined and analysed, which makes it hard to develop such algorithms for all problem variants. Finally, heuristic methods usually use some well designed rules to construct the schedules, however, they do not provide any guarantee on the quality of the obtained solutions. Because of that they are the most flexible and easiest to design. 

\begin{figure}[t]
	\includegraphics[width=\columnwidth]{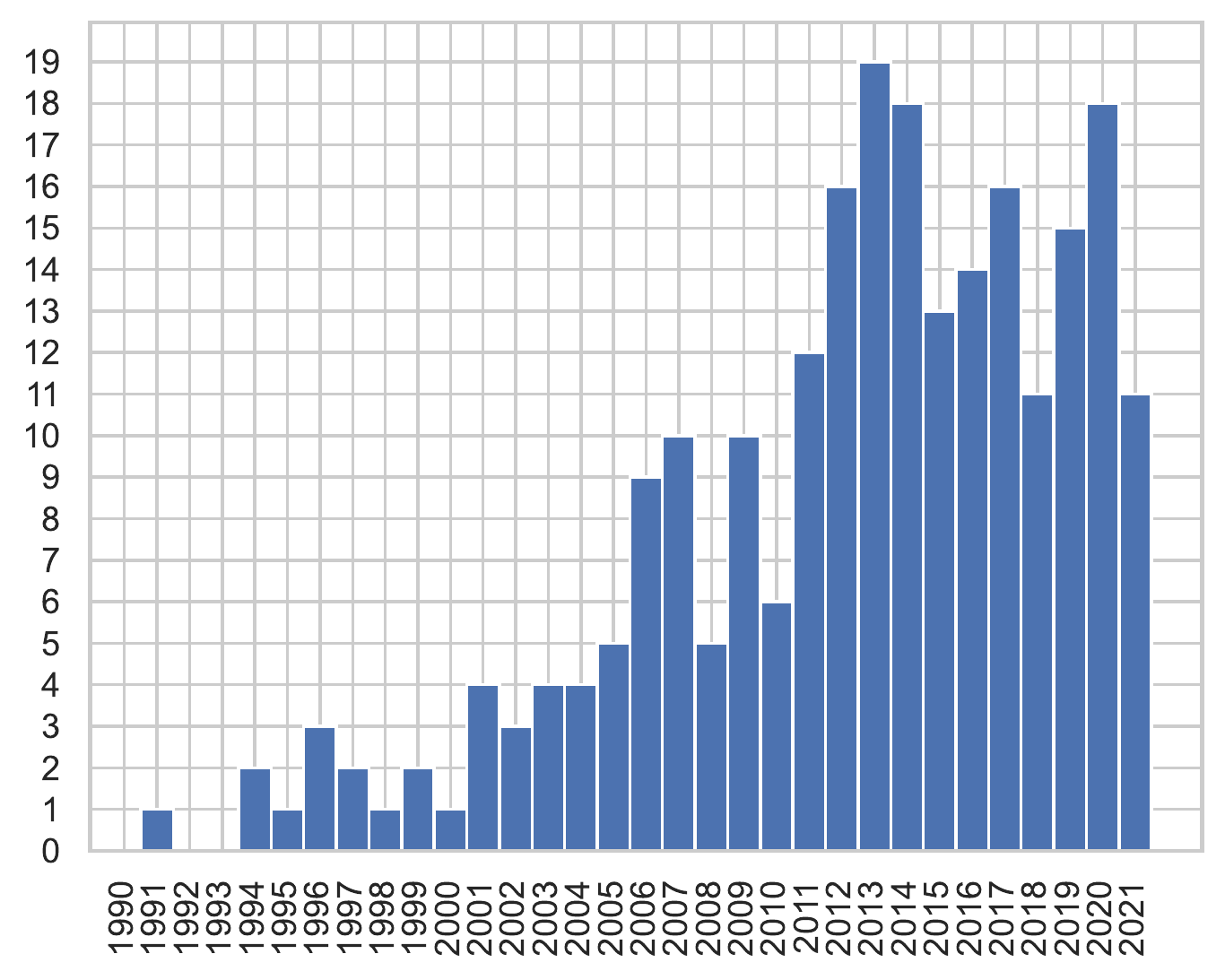}
	\centering
	\caption{Distribution of papers in the last 30 years}
	\label{fig:dist}
\end{figure}

Since there are many variants of scheduling problems, they have been categorised into different models and variants. For example, timetabling deals with scheduling students and teachers to events (lectures, exams) in schools and universities, rostering deals with scheduling employees to different shifts, job-shop deals with production environments in which a product needs to go through a series of steps until it is completed, and parallel machine environment deals with problems where each activity can be processed by multiple resources. The parallel machine environment can be further divided into three categories: identical, uniform, and unrelated. The unrelated parallel machines scheduling problem (UPMSP) is the most general from the parallel environments, as it makes the least assumptions regarding the resources that process activities. Although this environment has applications in many areas like, computer multiprocessor task scheduling \cite{WU2018}, equipment scheduling \cite{GEDIK2018}, and manufacturing \cite{Yu2002}, it received less attention than some other environments. However, recent years have seen the rise of new research dealing with this problem. Therefore, the number of problem variants that were considered in the literature and the solution methods applied for solving them has grown significantly. This is especially true for the solution methods, where the earlier research was more focused on exact and approximate methods. However, recent years have seen the rise in application of heuristic and metaheuristic methods. This can best be seen from Figure \ref{fig:dist} which shows the distribution of papers in which heuristics are used for solving the UPMSP over the last 30 years. The last 10 years show a rising trend in the number of studies which apply heuristic methods for solving the UPMSP.

Because of the large number of problem variants and solution methods, several papers provide an overview of the literature dealing with scheduling problems. For example, an overview of staff scheduling and rostering was provided in \cite{ERNST2004}. Timetabling problems have been surveyed in \cite{BABAEI2015}. Studies dealing with flexible job shop have been surveyed in \cite{Chaudhry2015}. An older, but detailed survey of classical scheduling problems has been provided in \cite{Chen1998}. Other papers have reviewed problems with specific properties, like setup times \cite{ALLAHVERDI1999, ALLAHVERDI2008, ALLAHVERDI2015} or no-wait in process conditions \cite{ALLAHVERDI2016}. Others reviewed different solution methods used for solving scheduling problems like evolutionary algorithms \cite{Hart2005} or multi-population heuristics \cite{LEI2020}. However, except for a few older surveys which deal with the parallel machines environment \cite{CHENG1990, Chen1998, Mokotoff2001}, the literature dealing with the UPMSP has not been surveyed in recent years. Because of the increase of research being performed in this area during the last years, it is becoming difficult to follow which research was performed, and what represents the current state of the art in this field. This increases the possibility similar research being conducted, or not considering the latest methods and results obtained by other researchers. 

The goal of this paper is to provide a thorough literature survey applying heuristic methods for the UPMSP. The research surveyed in this paper will be classified by two aspects, the methods applied for solving the problem, and the problem variant that was considered. As such, this survey provides an overview to easily find which methods have been applied on a considered problem variant, or all the problems on which a certain method has been applied until now. Such a systematised overview should help other researchers to obtain a thorough overview of UPMSPs, and hopefully broaden the research and lead it into new directions. 

The remainder of the paper is organised as follows. Section \ref{sec:desc} provides the description of the UPMSP. The literature overview is provided in Section \ref{sec:ovierview}. The reviewed papers are additionally classified into several categories in Section \ref{sec:class}. A overview of possible future directions is given in Section \ref{sec:outlook}. Finally, Section \ref{sec:conc} gives the conclusion of the paper. 

\section{Problem description and notation}
\label{sec:desc}

The UPMSP is defined by a set of $n$ jobs where each needs to be executed on one of the $m$ available machines. The subscript $i$ is used to refer to a certain machine, whereas the subscript $j$ refers to a job. The most general property which needs to be specified for this environment is the execution time of job $j$ on machine $i$, which is denoted as $p_{ij}$. In the UPMSP it is assumed that this value needs to be specified for all job-machine pairs, and that no machine relations exits from which the processing times can be inferred (for example that one machine is two times faster than another, so it require half the time to execute the jobs). This distinguishes it from other parallel machines environments, where the machines are either identical (with the same speeds) or uniform (have different speeds, but they are the same for all jobs). This makes the unrelated machines environment the most general in the group of parallel machines environments. Depending on the problem variant, other properties are also defined for jobs. In certain cases jobs need to finish until a given due date $d_j$. Additionally, in some problems not all jobs are equally important, therefore a weight $w_j$ is defined for each job which is then used when calculating the objective value of the schedule. When the schedule is constructed, the completion time $C_j$ can be calculated for each job, based on which several objective functions can be calculated.

To easily describe scheduling problem, the $\alpha|\beta|\gamma$ classification scheme is often used \cite{GRAHAM1979}. However, as the number of scheduling problems increased, the notation also had to be expanded to encompass all the problem variants and optimisation criteria. In this scheme, $\alpha$ represents the machine environment, which in the case of unrelated machines is denoted as $R$. The second field $\beta$ represents additional problem constraints and characteristics, and can include zero or more of them. These include
\begin{itemize}
	\item setup times ($s$) --  when switching from one job to another on a machine, a setup operation of a certain duration needs to be performed to prepare the machine for processing the next job. The length of this operation is given with $s_{ijk}$, where $i$ denotes the job which has finished executing, $j$ the job that should be executed next, and $k$ the machine on which the jobs are processed,
	\item release times ($r_j$) -- jobs are not available immediately at the start of the system, but are released into the system over time. Each job has a release time $r_j$ which denotes when the job becomes available.
	\item machine eligibility ($M_j$) -- each job can be executed on only a subset of machines
	\item precedence constraints ($prec$) -- jobs are not independent and cannot be executed in an arbitrary order. A job $j$ can have several predecessors, all of which have to be finished before job $j$ can start executing. 
	\item batch scheduling ($batch$) -- jobs are grouped into batches which are scheduled. Two variants exist, serial ($s-batch$) and parallel batch ($p-batch$). In serial batch all jobs are executed one after another, however, usually there is no setup required between the jobs of the same batch. In parallel batch, jobs are executed in parallel on a machine, and the execution time of the batch is equal to the longest execution time of any job in the batch. For $p-batch$ there are many variants, where jobs can have different sizes, machines different capacities, and similar. 
	\item job sizes ($s_j$) -- concerned with batch scheduling, denotes that not all jobs take the same space in the batch
	\item machine capacities ($Q_k$) -- concerned with batch scheduling, denotes that the capacity of the batch depends on the machine on which it is executed
	\item deadline ($\bar{d}$) -- jobs are required to finish until a given time
	\item common due date ($d_j=D$) and common deadline ($\bar{d}=D$) -- all jobs have the same common due date or deadline until when they need to execute
	\item machine availability constraints ($brkdwn$) -- also sometimes referred to as breakdowns. Defines that machines are not available all the time, be it due to expected or unexpected situations
	\item auxiliary resources ($R$) -- additional resources required during machine setup or job execution. They can be in the form of workers which need to perform some adaptations, or different renewable or non renewable resources.
	\item changing processing time ($p_c$) -- processing times are adaptable and can change during time. They increase either due to job deterioration, or decrease due to learning effects or by using additional resources
	\item dedicated machines ($M_{ded}$) -- jobs have machines which are dedicated for them, meaning that they are executed faster on those machines than on other machines 
	\item rework processes ($rwrk$) -- after execution it is possible that jobs are faulty and have to be reworked again on one of the machines
	\item machine deterioration ($M_d$) -- machines deteriorate over time and the execution of jobs becomes slower or their capacity is lower
	\item machine maintenance ($M_m$) -- maintenances have to be allocated to machines either to keep the system at a desired level (preventive maintenance) or to fix broken machines (corrective maintenance)
	\item machine speed ($M_s$) -- machines can be executed with different speeds to process jobs faster
	\item load ($L$) -- jobs need to be loaded and transported to machines using a vehicle with a constrained capacity
\end{itemize}

The $\gamma$ field represents the criteria that are optimised. This field must have at least one entry, but can also have more in the case of multi-objective optimisation. The optimisation criteria include:
\begin{itemize}
	\item makespan ($C_{max}$) -- represents the latest completion time of a job $C_{max}=\max_j{C_j}$
	\item total weighted flowtime ($Fwt$) -- represents the total weighted time each job spent in the system $Fwt=\sum w_jF_j$, where $F_j=C_j-r_j$. A special case is the total flowtime $Ft=\sum F_j$, where all weights are equal to 1. This criterion is equivalent to the total weighted completion time $Cwt=\sum w_jC_j$
	\item maximum flowtime ($F_{max}$) -- the maximum flowtime value of all jobs $F_{max}=\max_j{F_j}$
	\item total weighted tardiness ($Twt$) -- represents the sum of weighted tardiness of all jobs $Twt=\sum w_jT_j$, where $T_j=\max(C_j-d_j,0)$. A special case is the total tardiness ($Tt$) when all the job weights are equal to 1.
	\item maximum tardiness ($T_{max}$) -- maximum tardiness value of any job $T_{max}=\max_j(T_j)$
	\item weighted number of tardy jobs ($Uwt$) -- $Uwt = \sum w_j U_j$, where $U_j$ equals 1 if a job finished after its due date. A special case is the number of tardy jobs ($Ut$) when all weights are equal to 1.  
 	\item total energy consumption ($TEC$) -- represents the total energy consumed during system execution. Concrete definitions of this criterion can vary
	\item total weighted earliness and tardiness ($Etwt$) -- represents the sum of the weighted earliness and tardiness values for each job $Etwt=\sum w_j\max(d_j-C-j,0)+w_j(C_j-d_j,0)$. The weight for the earliness and tardiness part can be equal or different
	\item machine load ($ML$) -- considers the load balance across all machines. Usually as the difference of the total execution time between the highest loaded and lowest loaded machines.
	\item Cost ($COST$) -- the production cost of certain parts of the system execution. The definitions of this criterion can vary. 
	\item total setup time ($Ts$) -- total time spent on setups
	\item Total resources used ($R_u$) -- total amount of resources used in the schedule
	\item Total late time ($L$) -- total late time of all jobs, which is calculated in the same way as tardiness except if a job started executing after its due date, then the penalty is fixed regardless when it started.
	\item Number of jobs finished just in time ($N_{jit}$) -- number of jobs that finished exactly at their due dates.
\end{itemize}

\section{Heuristic methods for the unrelated machines environment}
\label{sec:ovierview}


The methods for solving the scheduling problem can be roughly divided into exact \cite{FANJULPEYRO2020}, approximate \cite{Lenstra1990}, and heuristic methods. A good overview of the former two methods is given in \cite{Wotzlaw2007}. The rest of this section will provide a detailed review on heuristic methods used for solving the UPMSP. These methods are divided into problem specific heuristics and metaheuristics. The former group consists of methods specifically designed for solving the UPMSP, whereas the second group consists of general heuristics that are adapted for solving the considered problem. Problem specific heuristic methods will be reviewed in Section \ref{sec:psh}, which are classified either as DRs which are reviewed in Section \ref{sec:dr}, and general heuristics reviewed in Section \ref{sec:gh}. A review of metaheuristic methods is provided in Section \ref{sec:meta}. Due to a large number of methods and terms that appear in the text, a list of abbreviations is given in Table \ref{tbl:abbr}.

\begin{table}[]
	\centering
	\caption{Abbreviations used in the text}
	\label{tbl:abbr}
	\begin{tabular}{@{}ll@{}}
		\toprule
		Term                                          & Abbreviation \\ \midrule
		Ant conony   optimisation                     & ACO     \\
		Apparent tardiness cost                       & ATC     \\
		Apparent tardiness cost with setup times      & ATCS    \\
		Artifical bee colony                          & ABC     \\
		Automatically designed dispatching rule       & ADDR    \\
		Branch and bound                              & B\&B    \\
		Cat swarm optimisation                        & CSO     \\
		Clonal selection algorithm                    & CLONALG \\
		Combinatorial evolutionary algorithm          & CEA     \\
		Differential evolution                        & DE      \\
		Dispatching rule                              & DR      \\
		Eearliest completion time                     & ETC     \\
		Earliest due date                             & EDD     \\
		Electromagnetism-like algorithm               & EMA     \\
		Estimation of distribution algorithm          & EDA     \\
		Evolution strategy                            & ES      \\
		Fixed set search                              & FSS     \\
		Fruitfly algorithm                            & FA      \\
		Greedy randomized adaptive search   procedure & GRASP   \\
		Genetic algorithm                             & GA      \\
		Genetic programming                           & GP      \\
		Genetic simulated annealing                   & GSA     \\
		Harmony search                                & HS      \\
		Harris hawk optimisaiton                      & HHA     \\
		Imperialist competitive algorithm             & ICA     \\
		Inteligent water drop algorithm               & IWDA    \\
		Iterative descent                             & ID      \\
		Iterated greedy                               & IG      \\
		Iterative local search                        & ILS     \\
		Learning automata                             & LA      \\
		Local search                                  & LS      \\
		Longest processing time                       & LPT     \\
		Manually designed dispatching rule            & MDDR    \\
		Multi-agent                                   & MA      \\
		Multi-objectibe                               & MO      \\
		Non dominated sorting genetic algorithm   II  & NSGA-II \\
		Particle swarm optimisation                   & PSO     \\
		Path relinking                                & PR      \\
		Record to record travel                       & RRT     \\
		Salp swarm optimisation                       & SSA     \\
		Scatter search                                & SS      \\
		Shortest processing time                      & SPT     \\
		Simulated annealing                           & SA      \\
		Sine-Cosine Algorithm                         & SCA     \\
		Squeky wehll optimisation                     & SWO     \\
		strength pareto evolutionary algorithm        & SPEA2   \\
		Tabu search                                   & TS      \\
		Treshold acceptance                           & TA      \\
		Variable neighbourhood descent                & VND     \\
		Variable neighbourhood search                 & VNS     \\
		Weighted shortest processing time             & WSPT    \\
		Whale optimisation algorithm                  & WOA     \\
		Worm optimisation                             & WO      \\ \bottomrule
	\end{tabular}
\end{table}

\subsection{Problem specific heuristics}
\label{sec:psh}

\subsubsection{Dispatching rules}
\label{sec:dr}
In the context of scheduling problems, a special kind of heuristics, denoted as DRs, were often applied \cite{Panwalkar1977,morton1993}. These heuristics create the schedule incrementally by assigning jobs on free machines by ranking them using a priority function. Based on their rank, the DR selects which job should be scheduled next. The advantage of such heuristics is that they are simple and fast, which makes them applicable for large problems. However, the downside is that they usually cannot match the performance of more complex heuristics.

The first study in which DRs are proposed for the UPMSP is \cite{A165_Ibarra1977}. In this study, 5 DRs based on the LPT rule, which schedules jobs with the longest processing time first, are adapted for minimising the makespan. These include the later commonly used min-min and max-min DRs. This research is further expanded in \cite{A38_De1980}, where the previous 5 DRs are analysed based on which a new DRs is proposed. This DR first approximates the makespan by another rule. When certain machines reach a load equal to the makespan multiplied by a certain factor, it does not consider these machine in further scheduling decisions. In that way, the DR distributes the load evenly to all machines.

In \cite{A80_HARIRI1991} a two phase method LP/ECT is proposed for minimising the makespan, which applies LP to construct a partial schedule and then the ECT rule to schedule the remaining jobs. Experimental results show that such a method achieves a better performance than using ECT by itself, but with a larger execution time. The authors further apply improvement procedures to all methods and show that the results of ECT can be improved significantly and match the performance of LP/ECT. The authors conclude that this makes ECT with improvement procedures a strong contender to be used in practical problems. In \cite{A129_Randhawa1995} the authors examine how different design choices in DRs affect their performance. Decisions like ranking jobs on machines (using LPT or the EDD rules), assigning jobs to machines and similar were considered. Setup times were defined for jobs and three objectives were optimised, the flowtime, number of tardy jobs, and machine load. Based on the experiments, the authors outline the important design choices for each objective. 

An extensive comparison of DRs for minimising the makespan was performed in \cite{A207_MAHESWARAN1999}. The paper considers the heterogeneous computing environment, which is identical to the $R||C_{max}$ problem. The authors compare 5 existing and propose 3 novel DRs. The experiments are performed on 4 sets of problem types that differ in machine and job heterogeneity, which specifies the difference in magnitudes between the processing times of jobs across the machines. The results demonstrate that the proposed sufferage rule is superior to other rules. In \cite{A245_Braun1999} and \cite{A18_BRAUN2001}, the authors perform a detailed comparison between 11 methods for minimising the makespan. The tested methods include 5 DRs, a GA, SA, GSA, TS, and A*. All algorithms were examined on problems with different heterogeneity properties. The results demonstrate that GA achieved the best performance, closely followed by the min-min rule. 

In \cite{A93_WENG2001} the minimisation of the total weighted flowtime with setup times was considered. The authors propose 7 DRs based on the WSPT rule for the single machine environment. They demonstrate that the DR which orders jobs based on the ratio between the processing time plus setup times and the job weight performs best among the tested rules. A novel DR is proposed in \cite{A243_Wu2001} for minimising the makespan. The authors provide an analysis of the existing min-min rule, and based on which they propose the relative cost (RC) rule. This rule takes into account both processing times and completion times of jobs on all machines and balances between the two measures to schedule the next job. A novel DR called minimum execution completion time (MECT) for makespan minimisation is proposed in \cite{A61_Kim2004}. This rule represents a combination of two simple DRs and alternatively uses the execution and completion times when scheduling jobs. It selects jobs by their processing times if this does not increase the makespan, otherwise it uses the completion times of jobs to select the next one. In \cite{A250_GolcondaDO04} a DR for optimising the number of tasks which meet their due dates is proposed. The DR sorts the jobs in order of their due dates and assigns them to the machine with the smallest completion time. If the job cannot meet its due date it is discarded and not scheduled at all. 

Four DRs for the batch scheduling problem with flowtime minimisation are examined in \cite{A192_Arnaout2005}. A novel DR which uses two priority functions is proposed, one for ordering jobs and the second to assign them to machines. The proposed DR exhibited a better performance than other applied DRs. The previous research is extended in \cite{A138_Arnaout2006} by considering stochastic execution and setup times. In these cases the real processing times are not known until the job starts executing on the machine. A batch scheduling problem with the goal of minimising the total weighted tardiness is considered in \cite{A130_Na2006}. In this problem, a batch is not allocated to only a single machine, but rather a set of machines that can process the jobs contained in the batch simultaneously. The authors apply standard DRs, DRs adapted for batch scheduling (which first sequence the batches, and then allocate the batches to the corresponding machines), and SA. SA achieved the best performance, closely followed by the DRs adapted fo batch scheduling.  

An interesting method of using reinforcement learning for solving scheduling problems is proposed in \cite{A121_Zhang2006}. The scheduling problem is modelled as semi-Markov decision process. Five DRs are used as actions during the learning phase and the reward function is based on the minimisation of tardiness. A Q-learning algorithm is applied to solve the defined reinforcement learning problem and compared to individual DRs. In all cases Q-learning significantly outperformed all the tested DRs and demonstrated that it can select good actions at various decision points. A comparison between 5 DRs is performed in \cite{A212_Xhafa2007}. The authors evaluate their performance on four criteria: makespan, flowtime, machine utilisation, and matching proximity (defines how many jobs are scheduled on jobs which execute them the fastest). The rules are tested under different heterogeneity conditions. The results show that there is no single rule that performs well over all criteria, however, the min-min DR achieved the best results for optimising the makespan and flowtime. A set of 20 DRs, out of which 17 are novel, is analysed in \cite{A248_LUO2007}. All newly proposed DRs follow the same structure. They first select the best machine for each job based on one rule, and then among all the pairs select the job that will be scheduled on its selected machine using a second rule. The authors propose integrating a task priority graph based on the consistency between jobs and machines into the rules. 

A novel DR called MaxStd is proposed in \cite{A29_Munir2008}. The goal of this DR is to prioritise jobs with a high standard deviation of their processing times, since these jobs could suffer the most if not scheduled on the right machine. The DR is applied for minimising the makespan and is compared to 5 existing DRs. In \cite{A19_TSENG2009} the authors compare the performance of 5 DRs when considering setup times and three objectives, the makespan, total weighted tardiness, and computing cost. The authors combine the ATCS rule with the minimum completion time strategy to improve its performance. The results show that the proposed DR achieved equally good results as other rules across the optimised criteria. 

In \cite{A54_Chen2009} the authors propose a DR for problems with setup times and machine eligibilities. The objective is to primarily minimise the weighted number of tardy jobs, and secondly the makespan. In this problem it is considered that recipes can be attached to machines, and that the setup times between jobs of the same recipe do not exist. The proposed DR uses different job assignment mechanisms depending on the number of waiting jobs and available machines. The DR is further improved by a LS with three neighbourhood operators. A novel DR called min-max is proposed in \cite{A208_Izakian2009}. This rule adapts the min-min heuristic, but it selects jobs based on the ratio of their minimum execution time and the execution time on the selected machine. The intuition behind this strategy is that the job is scheduled on a machine on which it can execute quickly. The rules is tested for optimising the makespan and total flowtime, and shows a better performance than other existing DRs. 

A new DR for scheduling problems with precedence constraints and makespan minimisation is proposed in \cite{A97_Liu2011}. This DR prioritises the jobs with a larger deviation of their processing times, whereas the machines are prioritised by the speed by which they process the jobs on machines. A scheduling problem which includes setup times and resource constraints is examined in \cite{A126_Ruiz2011}. Resources are used when performing setups and their use needs to be minimised, since they introduce an additional cost. The authors optimise the sum of the total flowtime and the amount of resources used for setups. Several DRs are proposed for solving this problem, which prioritise jobs with lower processing and setup times. 

In \cite{A17_RAMEZANIAN2012} the authors consider a scheduling problem with rework processes. This information is probabilistic, which means that only after a job is processed will it be known whether the job has to be reprocessed or not. Because of this, it is difficult to apply methods that search the entire space, and therefore the authors propose five DRs for minimising the makespan. An extension of the sufferage DR is proposed in \cite{A30_Rafsanjani2012}. This rule works in the same way as sufferage, however, it scales the priority values of jobs with a quotient between the processing time and completion times of jobs. In that way it can better judge whether the machine is appropriate for executing a job. The rule is compared to several others and shows its superiority for makespan and flowtime minimisation. 
 
An analysis of 6 DRs from the literature is performed in \cite{A249_Briceo2012}. The authors propose an iterative procedure to minimise the total execution time of jobs on non-makespan machines. This is done by creating an initial schedule, removing the makespan machine and all jobs allocated to it from the problem, and trying to create a new schedule with this reduced problem. The provided analysis shows that such a procedure can decrease the execution time on non makespan machines. In \cite{A90_YangKuei2013} the authors perform an analysis of several DRS for scheduling jobs with release times and optimising the makespan, weighted flowtime, and weighted tardiness criteria. For each criterion the authors propose a DR which is further improved by a LS executed after the DR. A weighted combination of the makespan and number of tardy jobs criteria is minimised in \cite{A114_Wang2012}. The considered problem included job release times, machine eligibility constraints, and setup times. The authors propose a DR based on EDD, which schedules jobs to reduce the setup times and also prioritises those jobs that could become late. After the DR constructs the schedule, three LS operators are applied to further improve the obtained result. 

A parallel batch scheduling problem is considered in \cite{A166_LI2013}. The authors propose two kinds of DRs to solve the considered problem. The first kind schedules jobs to batches, and  allocates the batches on machines. The second group of DRs first schedules jobs to machines, and then constructs batches out of the scheduled jobs. The results show that the heuristic which first allocates jobs to machines achieves the best results. In \cite{A110_Kuei2014} a problem with setup times, release times, and the total weighted tardiness objective is considered. The authors proposed a novel DR called ATCSR\_Rm which takes machines into consideration when calculating the priority values of jobs. The authors also apply EMA and integrate several LS operators into it. In \cite{A148_Strohhecker2016} the authors consider the problem of minimising the total flowtime and total setup time. The authors test several DRs in different scenarios (with varying processing time and setup time speeds) to determine their behaviour in different situations. The consideration is restricted only to two machines. They demonstrate that the best DR strategies obtain results close to optimal results and can improve the overall production performance.

In \cite{A70_Santos} the authors present a new DR called OMCT for makespan minimisation. In this rule the jobs are sorted according to a priority calculated based on their sufferage value and standard deviation of processing times. Each job is then scheduled on the machine on which it will complete the soonest. The problem of minimising the energy and tardiness cost is considered in \cite{A185_Li2015}. In this problem variant machines have three operating modes (operation, wait, stop) with different energy consumptions. The authors propose several DRs and two heuristic algorithms based on the simple DRs. A parallel batch scheduling problem is considered in \cite{A40_ARROYO2017} with the objective of minimising the makespan. In the considered problem jobs have different sizes, therefore not all batches will consist out of the same number of jobs. The authors propose three DRs for the considered problem. The first two are based on existing rules, whereas the third is proposed in the paper and selects whether it is better to create a new batch, or add the current job to an already existing batch. 

An overview of 26 DRs (24 from the literature, and two novel ones) are examined in \cite{A1_DURASEVIC2018}. The methods were tested on a scheduling problem with job release times and for optimising 9 criteria. Four data sets with different job and machine heterogeneity properties were used for testing. A batch scheduling problem with releasee times and unequal job sizes is investigated in \cite{A213_Zarook2021}. Several simple DRs used for creating batches and scheduling them on machines are proposed. These DRs work in two ways, the first group constructs the batches and then allocates them to the machines, whereas the second group first allocates jobs to machines, and then groups them into batches. The authors also propose a genetic algorithm for the minimisation of the problem. 

Aside from manually designed DRs, the application of GP and similar methods for automatic development of DRs has become increasingly popular over the last several years \cite{Branke2016}, \cite{Nguyen2017}. The first application of GP to generate DRs for the unrelated machines environments was considered in \cite{A225_DURASEVIC2016}. GP is used to evolve a priority function used to rank jobs and machines when creating the schedule. The authors considered release times and four scheduling objectives, makespan, total flowtime, total weighted tardiness, and number of tardy jobs. The evolved DRs are compared with manually designed DRs and demonstrate a better performance. However, they still cannot match its performance of a GA. 

Automatically designing DRs for MO problems was considered in \cite{A231_Durasevi2017}. In this paper 9 scheduling criteria were considered in various combinations. The authors applied 4 MO algorithms: NSGA-II, NSGA-III, MOEA/D, HaD-MOEA. The automatically developed DRs achieved much better results than manually designed DRs. In \cite{A228_Durasevi2017} the authors propose the application of ensemble learning methods from machine learning to automatically designed DRs. The motivation for this research is that a single DR cannot perform well on all problem instances. Therefore, ensemble learning methods were adapted for this problem in order to create sets of DRs that perform their decisions jointly. In order to do that, four methods were tested, simple ensemble combination (SEC), BagGP, BoostGP, and cooperative coevolution. The obtained groups of DRs show a better performance in comparison with a single manually or automatically designed DR. In \cite{A233_Durasevi2019} the SEC method was further investigated in different scenarios to determine how its performs with different rules sets and ensemble construction methods. 

In \cite{A229_DURASEVIC2020} the authors examine different strategies which can be used to schedule jobs on machines in ADDRs. This includes the analysis of allowing idle times or not, as well as whether a single priority function should be used to determine the sequence and allocation of jobs to machines, or whether this decision should be split into two priority functions. The problem under consideration included release times and the total weighted tardiness was minimised. Previous studies on ADDRs focused on dynamic scheduling problems in which the decisions had to be performed on line during the execution of the system. However, in \cite{A227_Durasevi2020} the authors focused on problems in which the decision could be made prior to system execution. Four methods were proposed that can improve the performance of ADDRs in situations when all system information is available. The authors show that in these cases the ADDRs can match the performance of a GA, or achieve a better solution in a smaller amount of time. The objective was to minimise the total weighted tardiness with job release times. The automatic design of DRs for the total weighted tardiness criterion was considered in \cite{A226_JAKLINOVIC2021}. In this study different properties like release times, setup times, machine eligibility, precedence constraints, and machine unavailability periods were considered in various combinations. The ATC rule and a GA were adapted for solving all the combinations of these properties. The results demonstrate that ADDRs for most constraint combinations achieved a better performance than MDDRs, however, they could not match the performance of the GA. 

\subsubsection{General heuristics}
\label{sec:gh}
A heuristic for minimising the total tardiness is proposed in \cite{A103_SURESH1994}. The authors apply an algorithm to resequence all the jobs on the machines in the increasing order of their due dates. Additionally, improvement procedures (exchanging jobs) on the final solution are also applied to improve its performance. In \cite{A251_SURESH1996} the authors consider a scheduling problem with machine unavailability periods that can either be deterministic or stochastic. A heuristic method which takes into account the possible occurrence of machine unavailabilities during scheduling is proposed. If the heuristic is applied in a probabilistic scenario, it performs rescheduling procedures whenever it detects that a machine will not be available. A scheduling problem with precedence constraints and makespan minimisation is examined in \cite{A65_HERRMANN1997}. The authors propose several heuristic methods which prioritise jobs that could delay future tasks. In addition, SA is executed after these heuristics to improve the results. The obtained results show that the proposed heuristics obtain good, even optimal results in some cases. In \cite{A99_Randhawa1997} the authors devise a heuristic method to optimise three criteria (flowtime, total tardiness, and number of tardy jobs) considering several constraints like machine eligibility, setup times and product types. The proposed heuristic consist of several steps, where jobs are first grouped into tasks by product types to decrease setup times. Those tasks are then assigned to machines, and then individually sequenced on them. In the sequencing step of jobs several simple DRs rules are used, and the best for each optimised criterion is determined. Finally, the authors provide a detailed analysis on the effects of different system parameters on the results.

In \cite{A36_Bank2001} the authors consider the minimisation of the total weighted tardiness and earliness penalties with job releases and a common due date. The authors propose several constructive heuristics and iterative algorithms (SA, TA, iterative improvement, and multi-start heuristic) for solving the considered problems. The experiments conclude that the performance of constructive heuristics depends heavily on the characteristics of the problem, whereas among the metaheuristics, TA usually achieved the best results. In \cite{A57_DHAENENSFLIPO2001} the authors consider a scheduling problem with setup times. In this problem, setup operations and job execution introduces a certain cost that needs to be minimised. The objective is defined as a linear combination of the makespan and the total cost produced by setup and processing of jobs. The authors  define a heuristic procedure to solve the given problem, two methods for constructing initial solutions, and introduce an improvement phase to further enhance solutions. 

A heuristic for minimising the makespan with machine eligibility constraints is proposed in \cite{A195_Salem2002}. The heuristic assigns the jobs to machines in the first step based on their processing times. In the second step, the heuristic from \cite{A80_HARIRI1991} is adapted for considering machine eligibility constraints. The batch scheduling problem with setup times and total weighted tardiness minimisation is investigated in \cite{A98_KIM2003173}. The authors adapt two existing DRs, EWDD and SWPT, for this problem. SA and a problem specific heuristic are also proposed. The heuristic method first orders the batches using EWDD, and then allocates the batches to machines and fills them with jobs to minimise the weighted tardiness. The experimental results demonstrate that the DR methods achieved the worst results, and that SA significantly outperformed all other methods, which shows that metaheuristics perform better than heuristics specifically designed for the considered problem. 

In \cite{A49_Chen2004} a scheduling problem with setup times and additional resources is considered with the objective of minimising the makespan. The auxiliary resources are limited and a job cannot be scheduled if not enough resources are available. Resources can freely be attached to machines, but this process requires a certain setup time. The authors propose a heuristic method which is based on assigning jobs to machines with the smallest processing times, scheduling jobs that require the same resource one after another, and keeping the load balanced across all machines. In a comparison with SA the proposed heuristic demonstrated a superior performance. A problem from the textile industry represented as an UPMSP, is examined in \cite{A128_SILVA2006}. The characteristics of this problems are that jobs cannot be processed on all machines and setup times occur when two lots are interchanged. Therefore, the goal of this problem is to group the jobs into lots (or batches) that can be processed without invoking setup times. The authors define a specialised heuristic which first selects the job, then the batch into which it will be included, and finally the position in the batch. 

The problem of scheduling lots with setup times and machine eligibility is examined in \cite{A163_DOLGUI2009}. In this case, lots represent a serial batch of jobs between which no setup times are invoked. The authors propose a heuristic, based on the nearest neighbour strategy from the travelling salesman problem, which iteratively schedules the jobs to machines. In \cite{A108_LIN2011} the authors propose several heuristics for the individual optimisation of three objectives, makespan, total weighted flowtime, and total weighted tardiness. The first heuristic uses LP to construct the schedule and improves it with several neighbourhood procedures. The other heuristics use a DR to construct the initial solution, and improve the solution quality with LS. In \cite{A124_Xu2015} the authors consider a batch scheduling problem in which jobs can be split across different machines to improve the performance of the system. The jobs that are being processed require an additional quantity of certain resources that are available. The authors propose three heuristic algorithms based on several optimality properties, which reduce the original problem to several sub problems that are iteratively solved. 

In \cite{A211_POLYAKOVSKIY2014} a multi-agent system is developed for the optimisation of the total weighted earliness and tardiness criterion. The proposed system consists of three agent types with different fitness functions and roles that they need to achieve. Each agent uses different procedures (approximate and LS methods) to solve its own problem. A problem from Polyvinyl Chloride pipes is formulated as a scheduling problem in \cite{A182_LEE2014} and includes dedicated machines, setup times, and a common deadline for all machines. The objective is to minimise the total completion time of all jobs. The authors propose 3 heuristic procedures for assigning machines to jobs, and show that they outperform a baseline method by a significant margin.  

In \cite{A75_FanjulPeyro2017} the authors study a scheduling problem with renewable resources that are required for processing jobs and need to be assigned to machines. The authors propose the application of matheuristics, which represent a combination of mathematical programming and heuristics. In those heuristics a mathematical model is solved, however, some decisions are performed heuristically. A makespan minimisation problem with additional renewable resources is considered in \cite{A125_VILLA2018}. The authors propose several heuristics to tackle this problem. The first group of heuristics execute in three phases. In the first phase they order the jobs (based on different strategies), then construct the entire solution, and finally improve it by various LS operators. The second group of heuristics does not consider the resources at all, but rather constructs the mapping of jobs to machines. This is likely to produce infeasible schedules, and therefore a correction procedure is applied to fix the solution.

The energy conscious unrelated machines environment is examined in \cite{A162_CHE2017}. In this problem each job is additionally characterised by the electrical energy that is consumed when it is being executed on certain machines. The electricity prices are considered to change during the day. Therefore, in this model the time horizon is divided into several time periods, each with its electricity price. The goal is to minimise the total electricity consumption. The authors propose a two step heuristic in which the jobs are assigned to machines to minimise the total cost (with the property of being preemptive) and then in the second stage the jobs are scheduled without preemption using an insertion heuristic. A problem with setup times, machine eligibility constraints and total tardiness minimisation was considered in \cite{A158_PEREZGONZALEZ2019}. The authors first propose several heuristic methods that are adapted for the considered problem. These heuristics are based on ordering jobs by certain properties and iteratively inserting them in the schedule at the position which leads to the smallest value of the optimised criterion. The authors also apply CLONALG with GRASP and VND. 

\subsection{Metaheuristics}
\label{sec:meta}

One of the first applications of metaheuristic methods to the UPMSP was done in \cite{A2_GLASS1994} for minimising the makespan. The authors compare SA, TS, ID, and a GA. For the first three methods two neighbourhood strategies for inserting and switching jobs between the machines are proposed. Additionally, to improve the quality of the results, the MCT rule is used to create starting solutions for all methods. Since the results demonstrate that SA and TS are superior to GA, the authors propose a GDA which incorporates the neighbourhood structures of the other methods, and shows to perform equally well as other methods. An ID algorithm is proposed in \cite{A13_PIERSMA199611} to improve previous results. A greedy algorithm constructs the initial solution by assigning each job to the machine with the smallest processing time. Additionally, the authors improve the neighbourhood search from \cite{A2_GLASS1994} and achieve better results. In \cite{A104_SURESH1996} the authors consider a bi-objective problem where the makespan and maximum tardiness need to be optimised simultaneously. The TS method is adapted for this problem by keeping a set of nondominated solutions that are obtained during execution. TS is also applied in \cite{A102_Srivastava1998} for the minimisation of the makespan objective. The authors adapt TS with a hashing function to control the restriction list. The improved TS achieves better results in comparison to the standard method and a LP solution method. 

In \cite{A84_KIM2002223} a problem with lots and setup times for minimising the total tardiness objective is considered. In this problem, jobs represents lots which consists of several items that need to be processed. All items in the job belong to the same lot, which means that no setup time is incurred between them and that they all have the same processing time. The goal is to group items belonging to the same lot to reduce the setup times. A SA method is proposed, in which the neighbourhoods are generated considering both lots and items. The results show that using neighbourhood structures which work on jobs, rather than individual items, greatly improves the results.  A SA method for minimisation of the makespan with setup times is considered in \cite{A22_Anagnostopoulos2002}. Five neighbourhood operators are used for interchanging and inserting jobs on a single machine and between two machines. The proposed algorithm obtained optimal solutions on all problems which were of smaller sizes. 

A multi-population MO GA is proposed in \cite{A50_COCHRAN20031087} for optimising the makespan, total weighted completion time, and total weighted tardiness criteria. The idea of this algorithm is to define a weighted sum between those objectives and obtain initial solutions. These solutions are used to initialise the starting populations of a multi-population GA, where each population is evolved for optimising a single objective. The proposed method achieved a better performance than other MO algorithms at that time. In \cite{A88_PENG2004} the authors consider a problem with fuzzy processing times. Three fuzzy scheduling models are defined and a hybrid GA is proposed for solving the considered problems. The efficiency of the algorithm is analysed on several scenarios. A batch scheduling problem in which jobs can be split from the batches is considered in \cite{A252_LOGENDRAN2004}. In this problem the batches of jobs are already known, however, splitting those batches and rescheduling some jobs could improve the schedule. Additionally, job and machine release times are considered, as well as machine eligibility constraints. Four DRs are proposed, which are used to create initial solutions for TS. The authors propose several TS algorithms and show that for smaller and medium  sized instances the algorithm which focuses on exploitation performs better, whereas on the larger instances the TS algorithm which focuses on diversification achieves better results. 

In \cite{A246_Ritchie2003}, the authors combine the sufferage and min-min DRs with a LS procedure which is executed on the solutions that are obtained by these simple DRs. The results are compared with a GA and show that a combination of a DR with local search improves the results in comparison with a standard GA, even when the GA also uses the solution obtained by min-min in its initial population. A GA with sub-indexed partitioning genes is proposed in \cite{A180_JOU2005}. This means that the encoding uses special symbols in the solution to split the jobs that are scheduled to each of the machines. Several genetic operators are adapted for this representation. The method is used to minimise the total weighted earliness and tardiness together with machine utilisation. The problem of optimising the total weighted tardiness and machine holding costs was investigated in \cite{A79_CAO2005}. The holding cost is incurred when a machine is used for processing jobs. The goal in this study is to to minimise the number of machines that are used for executing jobs. The TS method is applied with three local search operators which apply job swaps and insertions.

A parallel GA is applied in \cite{A87_Gao2005} to minimise the makespan. This algorithm runs a standard GA on several computers on the same network using MPI. Unfortunately, the analysis is very vague and it is difficult to determine the benefits of such a method. In \cite{A159_Rojanasoonthon2005} a problem for optimising the total weighted number of tardy jobs with release times, deadlines, machine eligibility restrictions, and sequence dependant setup times. A GRASP method which consists of two phases is used. In the first phase a feasible solution is constructed by ranking all feasible jobs at each decision point by a greedy function, based on which a job is selected and scheduled. In the second phase a neighbourhood search is performed to refine the solutions obtained in the first phase. A detailed analysis of the entire algorithm is performed and experimental evaluation demonstrated it performs better than a dynamic programming approach. 

The problem of scheduling jobs with secondary resources and tardiness minimisation is considered in \cite{A21_CHEN2006}. It is presumed that these secondary resources are expensive and cannot be used to an arbitrary extent. In addition, setup times are associated to the attachment and deattachment of these resources to machines, and each machine cannot process all jobs. The authors propose a metaheuristic method based on the combination of TS, TA, and improvement algorithms. The method shows superior performance when compared to SA and the ATCS rule. A similar problem is considered in \cite{A53_Chen2006}, where the maximum tardiness is minimised. A metaheuristic procedure based on guided search and tabu lists is proposed and compared to EDD, and SA. The proposed algorithm significantly outperforms both of these procedures.

A scheduling problem with a common undetermined due date is analysed in \cite{A89_MIN2006}. The objective is to find a common due date for all jobs which minimises the weighted tardiness and earliness criterion. The authors propose 4 GA variants: a standard GA, combination of a GA with SA, combination of a GA with an improvement heuristic, and a combination of a GA, SA and the improvement heuristic. The results demonstrate that all variants achieve a similar performance. A TS method is applied for a problem with setup times and makespan minimisation in \cite{A25_Helal2006}. The authors generate the initial solution using the SPT rule, and define several perturbation operators for examining the neighbourhood of the current solution (by exchanging jobs on the same or different machine). The method is compared to an exact algorithm and the results show that TS underperformed for smaller problem instances, while on larger ones it achieved a better performance. 

The problem of minimising makespan with setup times is considered in \cite{A157_Rabadi2006}. The authors propose  Meta-RaPS, a novel metaheuristic which uses a constructive and improvement heuristics. In the construction phase, a simple heuristic is used to order the jobs and allocate them to the machines considering their processing and setup times. After that, a set of neighbourhood operators are applied to improve the solution by exchanging and inserting jobs. Thus, the procedure can be considered similar to ILS. A problem of minimising the makespan without additional constraints is considered in \cite{A188_GUO2007}. The authors apply SA and TS with a job swapping and insertion neighbourhood structures. In addition, they also apply SWO combined with an iterative improvement LS, as well as with SA and TS. The results demonstrate that the procedure which combined SWO with TS and ILS achieved the best results.

In \cite{A164_Kim2006} the authors consider the minimisation of the total tardiness objective in a problem with setup and job ready times. A TS is used for solving the considered problem. Three initial solution construction methods are tested, which are based on DRs to order the jobs and then schedule them on machines. Since a solution can have a huge neighbourhood that would need to be searched, two candidate list strategies are introduced to limit the neighbourhood. The first strategy considers only a single job per machine for swapping or inserting. In the second strategy all tardy or non tardy jobs (depending on the iteration) are considered for being exchanged. Better results are obtained by using the second strategy. 

VNS was applied in \cite{A59_dePaula2007} for a problem with setup times and minimising the sum of makespan and total weighted tardiness. The proposed algorithm uses an initial solution obtained using an adapted NEH procedure \cite{NAWAZ1983}, and applies three LS operators, for swapping jobs on one or two machines, and for inserting a job into a machine with the lowest makespan. Three GRASP versions with PR are also proposed. The experiments demonstrate that VNS produces better results compared to GRASP. In \cite{A86_Xu2007} the authors consider the minimisation of makespan in a batch scheduling problem. A GA is proposed which uses the random key encoding, meaning that the solution is encoded as a series of real numbers. The integer part of the number determines on which machine the job is executed, whereas the fractional part determines in which sequence jobs are executed. Additionally, when the jobs are scheduled and sequenced, they are grouped to form a batch of jobs that are executed in parallel. The results demonstrate that GA performed better than a commercial solver. 

One of the first applications of ACO for the minimisation of total weighted tardiness is done in \cite{A118_Zhou2007}. The algorithm uses two pheromone trails, one for selecting the machine on which to schedule the next job, and another for sequencing jobs. A LS operator is also integrated into the algorithm. A GRASP for a problem with setup times and for minimisation of a makespan and total tardiness sum is proposed in \cite{A137_Ravetti2007}. The algorithm constructs the solution by assigning jobs to machines on which they would finish the soonest, and then applies a LS. Additionally, PR is used to intensify the search, and several design choices in the GRASP are evaluated. TS and SA are applied in \cite{A171_Kim2007} to minimise the total tardiness for a problem which includes setup times and release times for jobs. For both methods the initial solutions are generated using 3 DRs, whereas for the improvement phase two neighbourhood operators are used by interchange and insertion of the jobs. The experimental results show that TS performed best when searching a larger neighbourhood similar as shown in \cite{A164_Kim2006}. 

A problem with setup times, machine eligibilities, and load balancing constraints for optimising the flowtime is studied in \cite{A198_Yildrim2007}. In this study a load balancing constraint is introduced which restricts the imbalance between all machines. The authors define a structure for a simple DR and use different strategies to select the next job and machine on which it should be scheduled. The authors also propose a GA which uses the aforementioned rules to generate the initial population. The results demonstrate that GA improved significantly the solutions obtained by the proposed DRs. In \cite{A56_Chen2008} the authors propose a hybrid metaheuristic which combines the concepts of VNS and TS. The method generates starting solutions by three DRs and applies 4 neighbourhood operators to improve the solution. The neighbourhood operators are applied in succession, meaning that the next one is used if the previous was unable to improve the solutions. The method is applied for a problem considering setup times with the objective of minimising the weighted number of tardy jobs. 

ACO was also applied in \cite{A94_Arnaout2008} for makespan minimisation with setup times, by solving it in two stages. In the first stage jobs are assigned to machines. Then in the second stage the sequence on these jobs is determined. For each of the two stages a different pheromone trail is used. A LS procedure is included in the algorithm to improve its performance. This research was further expanded in \cite{A20_Arnaout2009}. In this study the experiments were widened to include more problem instances, and the parameters were more thoroughly optimised. ACO is compared to three methods previously proposed in the literature, TS, Meta-RaPS and a partitioning heuristic. Although ACO achieved a better performance than the other algorithms, it was improved in a subsequent study \cite{A41_Arnaout2012}. This improvement uses a new pheromone update strategy which provides a better scaling than the previous one. The results show an improvement over the previous ACO version and other competitive algorithms. 

In \cite{A156_Raja2008} the problem of minimising the total weighted earliness and tardiness criterion with setup times is considered. The authors propose combining a GA with a fuzzy logic approach. The motivation for this approach comes from the fact that standard procedures had difficulties in handling such an objective. Therefore, the GA was applied for generating schedules when different weight combinations of the earliness and tardiness criteria were considered, whereas the fuzzy logic approach was used to select the best combination of weights. The proposed approach is compared to several standard GAs, and demonstrated its superiority. 

SA is applied in \cite{A52_Chen2008} for a problem with setup times and the objective of minimising the total tardiness. Additionally, some jobs have deadlines which must not be be broken. In the first step, ATCS is applied to generate an initial solution, which is further refined using two additional procedures. SA uses several neighbourhood operators to generate new solutions, some of which perform changes on single jobs, whereas others modify an entire chain of jobs. The proposed procedure improved the initial solution obtained by ATCS. In \cite{A77_VALLADA2011612} a problem with setup times and makespan minimisation is investigated. The authors apply a GA coupled with a crossover and mutation operators enhanced by LS. Additionally, the algorithm also includes a fast LS that is applied regardless of the genetic operators. The different GA variants are tested to determine the quality of each procedure and the proposed method obtains good results in comparison to others. 

A MO problem of minimising the total weighted earliness and tardiness with the makespan is examined in \cite{A78_Gao2008}. A novel GA is proposed to deal with this problem, which transforms the MO problem to a single objective problem using a weighted sum of objectives. However, the weights are self-adapting during the evolution process to ensure that neither criterion starts to dominate. A LS method is also introduced in the algorithm, and the entire algorithm is parallelised. In \cite{A83_Klemmt} the authors examine a batch scheduling problem with incompatible job families and release times, in which the total weighted tardiness objective is optimised. A variant of the ATC rule adapted for batch scheduling is proposed. A VNS algorithm, which uses several neighbourhood operators that work both with individual jobs and job batches, is also proposed. The method was compared to a mixed integer programming approach, and demonstrated its superiority. However, the authors outline that to take more constraints into account the extension is more difficult. 

In \cite{A147_TAVAKKOLIMOGHADDAM2009} a scheduling problem with precedence constraints, setup times, job release times with the objective of minimising the number of tardy jobs and flowtime is considered. A GA is proposed which performs the optimisation in two steps is proposed. First, the number of tardy jobs objective is optimised, and in the second phase the flowtime is optimised. However, in the second phase the objective value obtained in the first phase is used as a constraint that is considered by the GA. The results demonstrate that the GA obtains results only a few percent worse than the optimal solutions. In \cite{A170_Chavez} the authors optimise the total weighted flowtime. SA is applied to solve the considered problem and shows good efficiency in obtaining optimal results. Makespan minimisation is considered in \cite{A16_Charalambous}. In addition to the makespan, which is used as the main objective, two auxiliary objectives are used, total flowtime and number of machines which have their completion time equal to the makespan. The authors propose and apply a VND method that uses 5 neighbourhood operators and 3 neighbourhood search sizes. 

In \cite{A58_Chyu2010} the authors consider a bi-criteria optimisation problem of optimising the total weighted flowtime and total weighted tardiness with setup times. The authors apply a Pareto converging GA using two solution representations, a random key encoding and a list encoding scheme. The authors use two MO SA methods for comparison, and show that the GA achieves a better performance. In \cite{A67_Chyu2009} the same problem is considered. A competitive ES memetic algorithm, SPEA, and NSGA-II are applied. All methods use the random key encoding, and some improvements for the methods are proposed by combining them with a weighted bipartite matching method. The proposed evolution strategy achieved better results than other competitors since it lead to a better search of the solution space. 

A VND method with several neighbourhood structures is proposed in \cite{A72_FANJULPEYRO201055} for optimising the makespan criterion. The authors propose three neighbourhood structures and improve them further after initial experiments. The resulting method is simple but obtains extremely good results. AIS for the MO problem of optimising makespan and weighted tardiness and earliness is considered in \cite{A187_GAO2010}. Similar as in \cite{A78_Gao2008}, the problem is transformed into a single objective optimisation problem with self adapting weights for each criterion. The method uses a vector encoding where one list specifies the order of jobs, while the second list determines the allocation of jobs to machines. 

In \cite{A39_BALIN20116814} the authors apply a GA  with a matrix encoding for minimising makespan. This encoding represents solutions with a $n\times m$ matrix where each element denotes whether a job is scheduled on a machine or not. The authors also propose an adapted crossover and mutation operators for this representation. In \cite{A51_CHANG2011} the authors develop dominance properties when exchanging jobs on the same machine or between different machines, which are necessary conditions for an optimal solution. These properties are used to generate initial solutions for a GA and SA method for makespan minimisation with setup times. The results show that the GA which includes those properties outperforms a standard GA.  

The VND method from \cite{A72_FANJULPEYRO201055} is combined with a commercial solver in \cite{A73_FANJULPEYRO2011301}. Instead of providing only a two phase method where initial solutions are obtained by the solver and then improved by VND, the authors propose a slight modification. They solve a reduced problem, where only a few machine assignments are considered, and then those solutions are improved by VND. The results show that this method solved large problems almost optimally. In \cite{A74_FANJULPEYRO2012} the authors consider an extended problem in which the makespan minimisation is considered. The two problem variants that were examined are called not all jobs (NAJ) and not all machines (NAM). In these variants it is not required to schedule all jobs, or not all machines have to be used. The authors apply the methods from \cite{A73_FANJULPEYRO2011301}, where in the first phase the machines or jobs that should be considered for scheduling are selected (based on the simple ranking procedures), and then the VND algorithm creates the schedule for this reduced set. 

In \cite{A106_VALLADA2011} the authors apply a GA coupled with LS methods to minimise the makespan with setup times. The proposed GA uses an enhanced crossover operator based on the one point order crossover, which when creating a new solution uses a LS operator. In addition, the authors use a LS operator in the GA outside of the crossover operator. An IG procedure is proposed  in \cite{A107_Lin2010} for the optimisation of the total tardiness problem with setup times. This procedure first destructs the solution by randomly removing jobs from machines, and then in the next phase it inserts them to construct a complete solution. After that, a LS procedure is performed by exchanging the jobs on machines to further improve the solution. A bi-criteria problem of minimising the sum of the weighted flowtime and weighted tardiness is studied in \cite{A131_Mehravaran2011}. The authors apply TS which uses DRs to construct the initial solution. Several design choices of the algorithm are then investigated during experiments. In \cite{A209_Jolai2009} the authors consider the objective of maximising the weighted number of jobs that complete just in time. They propose a GA and a combination of a GA with LS approaches for inserting and swapping jobs. Additionally, the authors propose two constructive heuristic procedures to generate solutions for the problem. 

A scheduling problem with release times and for the minimisation of the makespan is solved using PSO in \cite{A11_Lin2013}. The authors use a representation with partitioning symbols to denote which jobs belong to which machine. Since the representation is discrete, the authors apply a discrete PSO variant which adapts the operators in velocity calculation. A LS procedure is also introduced to exchange jobs positions. In \cite{A43_Bilyk2010} problem adapted from a printed wiring board manufacturing environment is considered. In it jobs need to be transported into a storage space by a vehicle with a given maximum capacity. Also, multiple scheduling periods are considered. The authors apply VNS for the considered problem with 8 local search operators and an initial solution constructed by the ATC rule. In \cite{A254_Niu2011} a problem with setup times and makespan minimisation is examined. The authors apply CLONALG which includes a LS operator. Additionally, several algorithm parameters are made adaptive to enhance the search ability. 

A batch scheduling problem of minimising the sum of the total weighted flowtime and weighted tardiness is considered in \cite{A45_BOZORGIRAD2012}. In this problem job release times, machine unavailabilities, and machine eligibilities are considered. The authors use TS with different strategies that focus either on intensification or diversification of solutions during optimisation. A problem of minimising the weighted number of tardy jobs with release times and setup times is considered in \cite{A55_Chen2011}. The authors propose a novel metaheuristic, which can be considered a variant of ILS. The algorithm uses a set of neighbourhood operators that are iteratively exchanged, and are applied together with TS to skip already visited solutions. In addition, several dispatching rules are used to initialise the starting solutions. 

In \cite{A71_Fleszar2011} a multi-start VND procedure is proposed for minimising the makespan with setup times. The multi start algorithm makes several independent runs of the VND algorithm. In each run an initial solution is generated, and then a small and large neighbourhood searches are performed. They compare the proposed method with others and show that it produces significant improvements. The same problem is investigated in \cite{A82_HADDAD}, where the authors propose a novel metaheuristic named GARP, which is a combination of GA, VND, and PR. The goal of this algorithm is to apply crossover, mutation, LS (using simple job exchange operators) and the PR during the evolution each with a certain probability. Additionally, for the solution initialisation the authors use an adaptive DR. 

Restricted SA is applied for optimising the makespan criterion with setup times in \cite{A85_Ying2010}. This method restricts the choices considered during LS to eliminate moves which are ineffective. This method detects bottleneck machines and determines moves have the possibility of reducing the makespan. The experiments show that using such a restricted algorithm leads to better results than using standard SA. In \cite{A155_Rodriguez2012} the minimisation of the weighted flowtime criterion is considered. The authors use ABC for optimisation, which is enhanced by a LS mechanism and an IG method that destructs and reconstructs the solution. In \cite{A168_KESKINTURK2012} several methods for solving a problem with setup times and makespan minimisation are compared. The authors propose use ACO in which each node consists of subnodes that represent machines on which the job can be scheduled. The ants visit all nodes, but not each subnode, and out of this tour a feasible schedule is constructed. In addition, a GA and several DRs are applied. The proposed algorithm obtained the best results among all the tested methods. The problem of minimising the total weighted flowtime is examined in \cite{A177_Rodriguez2012}. The authors apply GRASP which uses the standard LS operators which  insert or swap jobs between machines, and combine it with PR. A special step called evolutionary PR, which performs the PR procedure on all solutions in the elite set, is also introduced into the algorithm.

A problem derived from the parallel robot scheduling system is examined in \cite{A196_akar2012}, which includes jobs release times and precedence constraints. The goal is to minimise the total tardiness criterion. GA and SA with initial solutions generated by DRs are applied. The authors propose a hybrid algorithm which combines those methods. In the hybrid algorithm, SA takes the best solution from the GA, improves it, and reinserts it back into the GA. Such a method demonstrated a better performance than individual algorithms. A resource constrained problem, in which limited human resources are available for performing setup tasks, is considered in \cite{A66_CAPPADONNA2013}. The objective is to minimise the makespan and GA is used for solving the considered problem. The same problem is also considered in \cite{A202_Capadonna2012}, however, the setup times depend on the selected job, machine and worker. The authors propose three GAs for the problem. The first two algorithms differ in the encoding schemes they use. The first uses a simple permutation scheme which specifies the jobs and then a decoding scheme assigns the workers and machines to those jobs. The second representation uses multiple arrays to represent the order of jobs, but also their allocations to machines and workers. The final GA combines these two, so the algorithm first uses the permutation scheme and then at one point switches to the multiple array encoding. The hybrid GA achieved a better performance than both individual variants. This research is further extend in \cite{A68_Costa2013} with a deeper experimental analysis. Additionally, the setup times in this study depend on the last executed job, selected job, machine and worker. 
 
In \cite{A48_Bitar2014} a problem from the photolithography workshop is modelled as an UPMSP, in which additional auxiliary resources exist. These resources need to be transported to machines and jobs that require the same resource cannot be executed on two machines in parallel. The goal is to minimise the total weighted flowtime. The authors use a GA enhanced with a LS procedure. The ACO algorithm is applied for the minimisation of the total weighted tardiness objective in \cite{A109_Lin2013}. The problem is solved in three steps, by first selecting a machine, after which a job is selected, and finally a machine is reselected. The algorithm uses two pheromone trails, one for machines and the other for jobs. In addition, a LS operator is also used to improve solutions. The authors also provide an improved ATC DR, which uses pairwise job interchanges to improve the final solution. A problem with precedence constraints and the total tardiness minimisation is considered in \cite{A135_Liu2013}. The authors first propose a simple DR based heuristic used to solve the problem. After that, the authors propose a hybrid GA with two subpopulations, one initialised with the DR based heuristic, and the second one initialised randomly. On these populations the standard genetic operators are then applied. 

Several MO problems, which consider the makespan, total weighted flowtime, and total weighted tardiness, are examined in \cite{A111_LIN2013}. The authors apply a GA with the idea of fitness ranking. The algorithm uses a permutation representation with partitioning symbols, a solution initialisation with a DR, and a LS operator that is performed during evolution. The authors also propose an ATC DRs for bi-criteria optimisation of total weighted flowtime and total weighted tardiness, where the scaling parameter in the DR is adjusted to put more focus on the flowtime or tardiness objectives. The problem of job splitting is again considered in \cite{A115_WANG2013}. The authors propose use DE coupled with LS to solve the problem of minimising the makespan. 

In \cite{A142_Low2013} the authors minimise the makespan criterion for a problem with non-renewable resources and machine eligibility constraints. Additionally, the processing time is controllable and depends on the amount of resources used. The authors apply two variants of ACO. The first algorithm simultaneously determines the assignment of jobs to machines and allocation of additional resources. The second algorithm first schedules all jobs by their default processing times, and then allocates the resources. A scheduling problem in which the processing times and due dates are not known with certainty , and are modelled as fuzzy variables, is considered in \cite{A145_TORABI2013}. Three objectives are minimised: makespan, total weighted tardiness, and total weighted flowtime. The authors propose the application of a MO PSO algorithm, which includes a concept of dominance to obtain a Pareto front of solutions.

A combined problem of production and distribution is considered in \cite{A169_Chang2014}. The production part is represented as an UPMSP. The objective is to optimise the total weighted flowtime. The authors use ACO to solve the combined problem. The algorithm first constructs the schedule, which is done by traversing the graph obtained by enumerating all mapping possibilities between jobs and machines. After that part, the algorithm solves the distribution part of the problem. A MO problem of minimising makespan, total tardiness cost and machine deterioration cost is examined in \cite{A172_BANDYOPADHYAY2013}. The problem was solved using NSGA-II, SPEA2 and a modified NSGA-II. The modified NSGA-II uses a mutation operator that is applied on the entire population and the goal of which is to divide the jobs evenly across all the machines. The modified NSGA-II algorithm demonstrated to find a better distribution of Pareto optimal solutions. 

A TS method, which uses 8 LS operators, is examined in \cite{A175_Lee2013} for minimising the total tardiness with setup times. In addition to standard operators that exchange a single job, the authors define several operators which insert or swap groups of jobs between machines. In \cite{A178_RODRIGUEZ2013} the authors consider the minimisation of the total flowtime and propose an IG algorithm. The authors use 8 construction schemes schemes, two strategies for solution destruction, and also propose three improvement strategies. Different parameter choices are evaluated on a problem set to determine the best possible algorithm variant. 

In \cite{A204_Chen2013} the authors consider a problem with setup times in which the total weighted flowtime is minimised. In this problems a certain number of jobs have fixed deadlines which must not be violated. Therefore, this group of jobs needs to be prioritised. The authors apply a RRT metaheuristic and a random descent search. A revised SWPT rule to generate initial solutions is also proposed. A hybrid metaheuristic is used in \cite{A206_Diana} to solve a problem with setup times and makespan minimisation. The authors use CLONALG with GRASP to initialise the starting population and VND as a hypermutation operator. The research is further extended in \cite{A69_DIANA201594} with additional experiments and a more detailed comparison with other approaches from the literature. 

A hybrid metaheuristic called AIRP is proposed in \cite{A27_Cota} to minimise makespan with setup times. This methods is based on combining properties of ILS, VND and PR. The initial solution is constructed by a DR, which is iteratively improved by ILS and then VND. In \cite{A42_Celano2008} the authors model a scheduling problem from cell manufacturing. In this model job batches are being scheduled and a limited workforce capacity is also considered. The objective was to optimise the makespan and total tardiness. SA was used to help decision makers to determine the best sequencing of jobs and allocations of machines and workers.

In \cite{A46_AvalosRosales2014} a scheduling problem with setup times and the makespan minimisation is examined. A multi-start procedure, which consists of a solution construction and improvement phase, is used. The procedure creates a new solution in each iteration on which VND is applied for improvement. The problem of minimising the sum of the makespan and total tardiness was studied in \cite{A47_Caniyilmaz2014}. The problem includes setup times and machine eligibility restrictions. The authors propose the application of ABC and GA with integrated LS operators. The results show that the ABC algorithm achieved a better performance than GA. 

A problem of minimising the sum of the makespan and weighed earliness and tardiness criterion is solved using IWDA in \cite{A181_Kayvanfar}. The problem includes setup times and machine eligibility constraints. The algorithms is coupled VNS with 3 LS operators, and the results show that the hybrid algorithm performs better than the standard IWDA. In \cite{A100_2015} the authors consider the problem of minimisation the weighted earliness and tardiness criterion with setup times. The authors apply a GA with SA integrated as a LS with two neighbourhood structures. 

In \cite{A112_LIN2014} the authors solve a scheduling problem with setup times and makespan minimisation using ABC, which the authors adapt to the problem. The representation with partitioning symbols and a neighbourhood procedure are used. The procedure first destructs the solution by removing a job from the machine with the highest makespan and then reinserting them in all positions until a complete neighbourhood is constructed. A GA for minimisation of makespan with setup times is considered in \cite{A123_Eroglu2014}. The proposed GA uses the real key encoding and is embedded with a LS operator. Since the LS operators are usually defined on a permutation representation, they are adapted for the real number coding applied in the GA.

A combination of a GA and TS with B\&B for makespan minimisation is examined in \cite{A127_SELS2015}. These methods are combined with LS operators that perform specific changes on the solution. Additionally, they are hybridised with a truncated B\&B procedure to accelerate the search process. The method examines each job machine assignment and changes it if required. The total flowtime objective is optimised by SS in \cite{A150_Siepak2014}. To justify the use of SS the authors define a simple ILS method which uses the LS operators from SS on its own. The results show that better solutions are achieved by SS, however, the ILS method achieves solutions in a much smaller time.

In \cite{A167_LIAO2014} the authors model a scheduling problem based on sequencing inbound trucks in multi-door cross docking systems. The objective of this problem is to minimise the makespan and includes setup and job release times. Five metaheuristic methods are proposed for solving the problem, which include three variants of ACO (which differ in the solution representation), a SA-TS hybrid, and a SA-DE hybrid. The results show that the ACO variant that uses two vectors for coding the solution achieves the best results. A scheduling problem with rework processes is considered in \cite{A173_RAMBOD2014}. This model includes setup actions and machine eligibility constrains with the goal of minimising the makespan. The authors apply a GA and two ABC algorithms. A solution representation, which encodes the expected rework processes (which are considered finite) and schedules them accordingly, is used. The results demonstrate that the ABC algorithms outperformed the GA. 

A hybrid GRASP is proposed for solving the problem with setup times and minimisation of the total weighted earliness and tardiness criterion in \cite{A201_DECMNOGUEIRA2014}. It uses a greedy solution construction procedures that balances between greedily assigning jobs to the machine with the best objective value and randomly distributing jobs across machines. Additionally, the algorithm also uses PR and an ILS as an improvement phase. In \cite{A222_Haddad2014} the authors propose a novel metaheuristic method denoted as AIV for a problem with setup times and makespan minimisation. The proposed algorithm combines ILS and VND. This is done by performing and ILS with VND as a LS operator. The initial solution in the method is created using a DR. A makespan minimisation problem with precedence constraints is examined in \cite{A189_Jabbar2014}. The authors apply a GA and compare it to a simple DR which schedules the job which can be completed the soonest, and show that the GA achieves a better performance. 

In \cite{A95_Liang2015} the authors consider the optimisation of the sum of the total weighted tardiness and energy consumption. In the considered problem machines consume a certain amount of energy when standing idle. Therefore, machines can be turned on and off, however, turning the machine on consumes a certain amount of energy, therefore it is required to determine in which cases it is better for the machine to remain idle, and when it is better to turn it off. The authors propose the application of ACO based on the ATC rule. Solution construction is performed in three steps, first machine selection, then job selection, and in the end machine reselection. A LS method is also embedded in the algorithm. In addition, the authors also adapt a GRASP that was proposed for the single machine environment. 

A MO problem in which the makespan, total weighted flowtime, and total weighted tardiness are optimised, is considered in \cite{A113_Lin2015}. The authors apply a MO multi-point SA method that uses a permutation representation with partitioning symbols. The initial solution is constructed by the modified ATC rule. The procedure uses the neighbourhood structures to generate new solutions and adds them to the set of non dominated solutions. The same problem is also considered in \cite{A136_Lin2016}. In this study the application of an iterated Pareto greedy algorithm for MO optimisation is proposed. This algorithm keeps a list of Pareto solutions on which it performs destruction and construction operators to obtain new solutions and add them to the set of Pareto optimal solutions. This procedure is enhanced with elements from TS restrict the possible moves in future iterations.  In \cite{A122_Zeidi2015} the authors propose an integrated metaheuristic for solving a problem with setup times and minimisation of the total tardiness criterion. The proposed metaheuristic is a GA which uses SA as a LS procedure that tries to improve a solution from the last generation of the GA. 

In \cite{A132_Ebrahimi2015} the authors consider a scheduling problem in which machines deteriorate over time and require that maintenance activities are scheduled on them. During a maintenance activity the machine cannot execute any jobs. The objective is to minimise the makespan. The authors propose the application of a GA that with an extended representation that allows scheduling maintenance tasks to machines. A GA is proposed in \cite{A133_Liao2016} to solve a problem with setup times and makespan minimisation. The authors propose a simple constructive heuristic based on the adjacent index value that considers the flexibility of jobs. A GA with LS that is applied after the mutation operator, is also used for the considered problem. A scheduling problem with setup times and machine eligibility with the objective of minimising the total flowtime is examined in \cite{A140_Joo2015}. The authors propose a GA coupled with DRs to obtain feasible solutions after genetic operators. The chromosome encodes only the ordering of jobs, and 3 DRs are used to schedule the jobs on the appropriate machines. A batch scheduling problem with setup times, machine eligibilities, job release, times and machine availabilities is considered in \cite{A152_Shahvari2015}. The sum of the weighted flowtime and total weighted tardiness is optimised. The authors apply a TS procedure with several solution initialisation techniques, and LS operators which insert or swap jobs between batches. 
 
In \cite{A160_Afzalirad2016} the authors consider the problem of minimising the total late work. This criterion is similar to the total tardiness, except that in this case it a job starts after its due date, it receives a constant penalty, which then does not depend on when it started. If only a part of the job is executed after the due date, then it is calculated in the same was as for tardiness. The considered problem includes setup times, precedence constraints, job release times, and machine eligibility restrictions. The authors propose a hybrid GA that uses the acceptance strategy of SA, so that children can advance to the next generation even if they are not better than their parents. A MO optimisation problem of minimising the makespan and total weighted tardiness is examined in \cite{A179_LIN2015}. The authors propose a heuristic procedure that combines the NEH algorithm with ATCS, which is adapted for solving a bi-criteria problem. In addition, the authors also use TS that is also extended bi-objective optimisation by keeping a list of non-dominated solutions.  

A problem with deteriorating job effects, where processing times of jobs increases after they are released into the system, is analysed in \cite{A186_SALEHIMIR2016}. The authors apply a GA, PSO, and a hybrid method which combines the previous two metaheuristics. The hybrid metaheuristic uses the crossover and mutation genetic operators, but also upon finding a better solution it tries to move all the solutions to that solution similar as in PSO. A problem with release times, precedence constraints, machine eligibility restrictions, setup times and additional resources is examined in \cite{A24_AFZALIRAD2016}. The goal is to optimise the makespan objective. The authors apply GA and AIS (which includes parts of an IG procedure) for the considered problem. In \cite{A33_Wang} the authors consider a problem of makespan minimisation. They use EDA embedded with an IG procedure. The authors derive certain properties about neighbourhood operations and use those information are used to build a probabilistic model in EDA to focus the search to promising areas. A parallel batch scheduling problem with unequal job sizes and machine capacities is considered in \cite{A35_ARROYO2017}. In the considered problem the makespan criterion is optimised. The authors propose an IG algorithm that uses a DR to initialise the starting solution. After that it performs a perturbation operator and a LS procedure to locate a better solution. The authors also apply a GA, SA, and ACO. 

A resource constrained problem, which considers that certain tools need to be loaded to machines, is investigated in \cite{A62_OZPEYNIRCI2016}. Each tool can be used by a single machine and has to be removed after processing. Tool switching times are negligible, and the tools never breakdown. The authors use TS and show it obtains optimal solutions for optimising makespan. Another resource constrained problem with makespan optimisation is considered in \cite{A119_ZHENG2016}. In this problem, a finite number of renewable resources is available. A job cannot be processed if at least one unit of a resource is not available during its entire execution phase. The authors propose the application of FA with an initial solution generation procedure based on two DRs. The work performed in \cite{A120_Zheng2018} also focuses on a resource constrained scheduling problem. However, this problem is extended to consider the total energy consumption objective together with the makespan. Although the machines are unrelated, they have different operating speeds. Increasing the speed lowers the processing times of jobs, however, it increases the energy consumption. An additional constraint is that machine speeds can be changed only before a job starts executing. The authors propose a collaborative MO FA enhanced with initial solution construction and three LS operators.

A comparison of 5 stochastic LS methods with 6 neighbourhood structures for the problem of makespan minimisation with setup times was performed in \cite{A139_Santos2016}. The examined methods are SA, ILS, and 2 hill climbing variants. They are compared to the results of the AIRP from a previous research. The results show that SA achieved superior performance to compared to other algorithms. A scheduling problem which considers delivery trucks with heterogeneous capacities which need to be scheduled to deliver jobs to customers is analysed in \cite{A141_JOO2017}. The objective is to minimise the makespan. The authors propose a single stage GA which uses a permutation array to represent the sequence of jobs. Based on that sequence DRs are used to assign jobs to machines. Another rule is used to group jobs into batches and assign them to delivery trucks for transportation. A MO problem of minimising the total weighted flowtime and total weighted tardiness is considered in \cite{A161_AFZALIRAD2017}. The problem includes job precedence constraints, machine eligibility constraints, setup and job release times. The authors propose a MO ACO algorithm which uses two pheromone trails for sequencing and scheduling of jobs similar as in \cite{A168_KESKINTURK2012}. The algorithm also embeds heuristic information based on job processing times for the selection of machines. The NSGA-II algorithm is also applied, and compared to the ACO algorithm which achieved better results.  

Similar as in \cite{A176_Low2016} the authors consider a problem with controllable processing times that can be decreased linearly by using an additional non-renewable resource. The goal is to minimise the makespan. The authors apply two ACO algorithm variants. In the first the goal is to simultaneously determine the assignment of jobs to machines and the amount of resources that are used, whereas in the second the entire schedule is constructed without any resource allocations, and then the processing times are compressed by allocating resources to machines. A study on methods for minimising four scheduling objectives has been performed in \cite{A230_Durasevic2016}. The considered criteria were the makespan, total flowtime, total weighted tardiness, and weighted number of tardy jobs. The authors apply 4 DRs and a GA using a permutation and floating point encoding. The authors also measure the time required for the GAs to achieve solutions of equal quality as DRs. The results show that when starting from a random solutions GAs require substantially more time to reach solutions of equal od better quality than DRs.   

A MO problem of minimising the total completion time and the weighted earliness and tardiness criterion is examined in \cite{A64_Zeidi2017}. The problem also includes release times, setup times and machine eligibility constraints. The authors apply two MO algorithms to solve the considered problem, namely NSGA-II and CENSGA. Minimisation of makespan with setup times is considered in \cite{A96_Tozzo2018}. The authors  apply a GA and VNS on the considered problem. The GA uses a simple constructive heuristic to generate the initial population. Additionally, the GA also uses three LS operators after the application of genetic operators. The VNS uses the same method to generate the initial solution as GA and the same three LS operators to examine the neighbourhood. 

A batch scheduling problem with the bi-criteria objective of minimising a linear combination of the total weighted tardiness and weighted flowtime is examined in \cite{A151_Shahvari2017}. Machine eligibility, job release times, machine availabilities, and setup times are also considered. Zhe authors also consider that the batches have lower bound on the size, meaning that batches below a certain size cannot be constructed. An extended TS algorithm which works at three levels is proposed. In the first level the algorithm joins jobs into batches, in the second level the sequence of batches and their allocations to machines are determined, and finally in the last level is tasked with sequencing jobs within the same batch. A comparison of different MO methods for simultaneous minimisation of the makespan, weighted tardiness and earliness, and purchasing cost is done in \cite{A153_SHAHIDIZADEH2017}. A parallel batch scheduling with release times, varied job sizes, and machine capacities is considered. The authors test the MO HS, ACO, and PSO variants, as well as the NSGA-II algorithm. An extensive analysis has been performed and which shows that the MO HS algorithm achieves the best performance. 

In \cite{A194_SHAHVARI2017} the authors consider a batch scheduling problem which includes additional resources in the form of operators that have different skill levels. The operators are required to perform setups and run the batches on machines. It is possible that the same operator cannot perform both for the same batch. Two objectives, the makespan and production cost, are minimised. The problem also includes machine and job release times, setup times, and machine eligibility constraints. The authors apply a hybrid MO PSO algorithm for the optimisation of this problem. The goal of it is to split the solution construction, so that the assignment of jobs on machines is determined by PSO, but then the allocation of batches to jobs and their sequencing is determined by an auxiliary heuristic. In \cite{A203_Abedi2017} the authors consider a problem with ageing effects of machines. This means that the processing time of jobs that are scheduled increases the longer the machine is running. Each machine can be subjected to a maintenance task to revert it back to its original state. The objective of this problem is to minimise the sum of the total weighted tardiness and earliness and maintenance costs. A GA with LS is applied, which uses a two part coding scheme where the first part determines the sequence of jobs on machines, whereas the second one determines when the maintenance should be performed. The ICA is also applied for this problem. 

The problem of minimising the total weighted earliness and tardiness with dedicated machines is considered in \cite{A205_CHENG2017}. The authors apply a GA and extend it with a distributed release time control (RTC) mechanism which plans the job sequences and machine allocations. The goal in this research is to define the release times of machines, so that the machines start processing jobs in a way that the jobs complete as close as possible to their due dates. In \cite{A210_MANUPATI2017} the authors examine the problem with setup times, job ready times, and auxiliary resources with the objective of minimising the makespan, total weighted flowtime, total weighted tardiness and machine load variation. The authors also consider that scheduling environments are usually uncertain, and therefore the processing times and due dates are modelled using fuzzy sets. A novel MO algorithm is proposed, which is based on the NSGA-II algorithm with integrated elements from the immune based metaheuristics. The algorithm is compared to standard NSGA-II and MO PSO. 

The problem of minimising makespan with setup times is examined in \cite{A221_Cota2017}. The authors propose an adaptive local neighbourhood search that is coupled with LA. The motivation for introducing LA is to learn the probabilities for applying the different LS operators that are used in ALNS. Therefore, the goal is that the LA learns which actions are good, and that this information is used to guide the LS. A parallel batch scheduling problem with the goal of optimising the makespan is examined in \cite{A184_LU2018}. The authors consider deteriorating job processing times and mandatory machine maintenance periods that have to be scheduled. After a maintenance period the execution time of jobs will be shorter. The authors propose a hybridisation between the ABC and TS, in which TS is executed in each iteration on several individuals in the population to improve them. The solution encoding specifies the assignment of jobs to machines, whereas the grouping and sequencing of jobs on machine is done by an additional procedure. In \cite{A253_DIANA2017} the authors propose a metaheuristic immune network optimisation (INO) combined with VNS. The authors show that the algorithms based on immune networks may have problems with exploration that lead to a premature convergence, and therefore they combine it with VNS to improve the chances of escaping local minima. The proposed approach is tested on a problem with makespan minimisation and setup times. 

In \cite{A240_LU2017} the authors model a scheduling problem of a heating process as an unrelated machines environment scheduling problem. In it, jobs are arranged and scheduled in batches. Additionally, instead of the standard processing time, all jobs have heating and soaking times associated to them. The objective is to minimise the total energy cost that is incurred from the energy consumed during the heating and soaking processes. The authors apply a standard GA, and an adapted GA which uses genetic operators that consider more information about the problem. A problem with setup times and makespan minimisation is optimised with SOS in \cite{A190_Ezugwu2018}. The algorithm is improved by using LS and heuristic for assigning jobs to machines so that only the sequence of jobs is encoded in the solution representation. The research is further extended in \cite{A31_EZUGWU2019}. The authors apply the concept of the adjusted processing times matrix (which aggregates the information about the processing and setup times) for solution representation. They also incorporate LS improvement strategies in their algorithm and improve the individual operators of the algorithm to make it suitable for the considered problem. The authors also combine their enhanced algorithm with SA for further enhancement.

A problem with setup times and preventive maintenances are considered in \cite{A26_AVALOSROSALES2018364}. In this study preventive maintenance periods are fixed, and jobs need to be allocated in a way that they do not overlap with them. The authors propose a multi-start algorithm which consists of a solution construction phase, and two improvement phases both which use the same operators but are applied in different ways. In \cite{A28_ZHOU2018} a batch scheduling problem with different machine capacities and job sizes are considered with the objective of minimising the total makespan. The authors apply a GA with random key encoding to determine the sequencing of jobs and their allocation to machines. During solution evaluation the jobs are  allocated to batches and sequenced on machines using a first fit heuristic adapted from the bin packing problem. 

In \cite{A32_Afzalirad2015} the authors consider a problem with resource constraints and makespan minimisation. They apply two GA variants. In the first variant, a three level solution encoding is used, in which the first part represents the sequence of jobs, the second their allocation to machines, and the third the priority of assigning additional resources to jobs. The second GA uses a two level encoding without the priorities of additional resources and applies a heuristic for resource allocation. The results suggest that the first encoding scheme which includes all the information performs better. The problem of total weighted tardiness with setup times is examined in \cite{A60_DIANA2018}. The authors propose a hybrid metaheuristic that combines ILS with VND in a way that VND is executed as a LS operator in each iteration of the algorithm. 

In \cite{A117_WANG2019} the authors minimise the weighted flowtime using an enhanced version of TS to tackle the problem. They adapt the k-opt strategy commonly used for the travelling salesman problem and vehicle routing problems. Additionally, the authors implement an IG algorithm and test several options for each of the algorithm phases. The methods are tested on datasets with up to several thousand jobs, and demonstrate that the proposed TS algorithm outperformed the IG algorithm. The MO problem of makespan and total energy consumption minimisation is studied in \cite{A143_WU2019}. The problem assumes that machines can process jobs at different speeds, however, all machines have the same speed at a certain level. The energy consumed depends on the processing speed that is utilised during job execution. A DE is applied for the considered problem. The solution representation does not encode the allocation of jobs to machines, rather this is done by a simple heuristic during evaluation. Additionally, the algorithm also uses several LS operators for swapping jobs on machines and adjusting machine speeds. Since the order of these operators can have an influence on the results, the authors apply a meta-Lamarckian learning strategy that learns and adjusts the order of in which the operators should be applied. 

An improved FA is applied in \cite{A12_Ezugwu2018} for solving a problem with setup times and makespan minimisation. The authors apply a transformation procedure to apply the continuous FA for the considered problem. Several LS operators are included in the algorithm to improve solutions. The MO problem of minimising the total tardiness and energy consumption is studied in \cite{A220_Pan2018}. In this problem each machine consumes a certain amount of energy per time. The authors apply ICA and add several improvements to the algorithm specific operators. A resource constrained problem with renewable resources is studied in \cite{A3_VALLADA2019}. The objective of the studied problem is to minimise the makespan. The authors propose two metaheuristics, an enriched SS and IG algorithm, in which a restricted LS is embedded. Additionally, the starting solution is constructed using a heuristic from a previous study \cite{A125_VILLA2018}.

In \cite{A9_Lei2020} the authors consider the makespan minimisation in a distributed UPMSP, in which machines are grouped in factories. However, as no additional constraints exist between machines in different factories, this problem is equal to the basic problem as all machines can be considered for scheduling any job. The authors apply an ICA extended with an additional memory structure that stores good solutions which are used later in algorithm operators. The algorithm also applies four neighbourhood operators and a global search operator. A hybrid metaheuristic for the problem with setup times and makespan minimisation is proposed in \cite{A10_breu2020}. This metaheuristic, called GIVP, is based on a combination of GA, ILS, VND, and PR. After the GA is executed, ILS, VND and PR are executed in a loop until the stopping criterion is not met. Additionally, the initial solution is generated using a DR. 

A scheduling problem with setup times and preventive maintenance is examine in \cite{A14_Wang2019}. The preventive maintenance periods occur every fixed time interval and at that point no jobs can be executed. The goal is to optimise the makespan and total tardiness. The authors apply ICA which includes a multi-elite guidance strategy and several LS operators and concepts from EDA. The algorithm is also adapted for MO optimisation. In \cite{A15_ALHARKAM} the authors consider a resource constrained scheduling problem with makespan minimisation. There are several resource types with a given amount, and each job requires a certain amount of resources to be executed. The authors apply a two stage metaheuristic which is a combination of VNS and SA. The algorithm starts with an initial solutions constructed by several DRs and iteratively applies a LS, however, it uses a SA acceptance test for the solution. 

In \cite{A44_BEKTUR2019} the authors consider the problem with setup times, machine eligibility restrictions and a common server. The idea of the server is that it performs the setup times for machines  and can perform the setup for only one job at a time. However, the setup of a job for a machine cannot be started until the machine is free. The goal is to optimise the total weighted tardiness criterion. The authors adapt the ATC DR for such a problem, a TS and two SA methods (with the only difference being if the initial solution was generated by ATCS or randomly). A batch scheduling problem with job release times, and unequal job sizes and machine capacities is examined in \cite{A193_Arroyo2019}. The objective is to minimise the total flowtime. The authors propose an IG algorithm, which uses several DRs to initialise the starting solution and then iteratively performs the destruction and construction of the schedule, as well as a local search to further improve the solutions. 

SCA is applied  for scheduling problems with setup times and makespan minimisation in \cite{A223_Jouhari2019}. The authors modify the SCA with SA in a way that in each iteration SA and then the SCA algorithm operators are used to update the current solution. This process is repeated for the entire population until the termination criterion is not achieved. In \cite{A232_VLASIC2019} the authors consider the effects of initialising a population of a GA with randomly or different kinds of DRs. The problem under consideration included job release times and the goal was to minimise the total weighted tardiness. Both, manually and automatically DRs were tested, and it was demonstrated that initialising populations with any kind of DRs significantly improved the performance and convergence speed of the GA. 

In \cite{A234_YEPESBORRERO2020112959} a scheduling problem with additional resources and setup times is investigated. The goal was to optimise the makespan. The authors first propose several heuristic algorithms which construct the initial solution and apply a repair procedure to ensures that the resource constraints are not violated. The authors also apply a GRASP procedure to solve the considered problem. The minimisation of the total weighted tardiness and total energy consumed is investigated in \cite{A235_SOLEIMANI2020}. The problem includes several properties like  setup times and deteriorating job times. In this context, the real processing time is calculated based on two concepts, job deterioration and learning effects. Job deterioration leads to a longer execution time as the time from release increases, whereas learning effects lead to a lower execution times the latter the job is executed. The authors apply three metaheuristics: CSO, ABC, and GA. 

A unified heuristic that can be applied for a wide class of scheduling problems that include release and setup times, and focus on the minimisation of the weighted earliness and tardiness, total weighted flowtime and total weighted tardiness is proposed in \cite{A244_Kramer2017}. The proposed method is based on multi start ILS that, depending on the kind of the problem, invokes VND with appropriate LS operators. The proposed method showed a good performance across all the considered problems and even other machine environments. In \cite{A6_TERZI2020} the authors consider a problem with setup times and makespan minimisation. The authors apply a hill climbing method which generates a new solution in each iteration and then the VND method with give neighbourhood operators.

In \cite{A7_Ark2019} the authors consider the problem of minimising the weighted tardiness and earliness criterion with a common due date. The authors tested three metaheuristics, ABC, GA, and SA to examine which kind of metaheuristic type (swarm intelligence, evolutionary algorithm, single solution algorithm) achieves the best results. The algorithms did not include any problem specific modifications, and the results demonstrated that ABC achieved the best performance. A MO optimisation problem of makespan and total tardiness minimisation is examined in \cite{A8_Lei2020}. An improved ABC method is presented which includes problem specific LS operators, and modifications to several algorithm operators. The application of WO for the problem with setup times and makespan minimisation is performed in \cite{A34_Arnaout2019}. The solution is constructed similarly as in ACO, by assigning jobs to machines in the first phases, whereas the sequence of those jobs is determined in the second phase. Both are done using pheromone trails. A study of different solution representations for GAs was performed in \cite{A224_VLASIC2020}. The authors compare 7 representations which are either permutation based or real number based. The representations were tested on four criteria (makespan, total flowtime, total weighted tardiness, weighted number of tardy jobs) and it was demonstrated that permutation based representations achieved the best results, with the best result being obtained by the representation that encodes the sequence of jobs and their allocation to machines is determined using a simple strategy. 

The HHO and SSO methods have been applied in \cite{A91_Jouhari2020} for the minimisation of the makespan. A modified HHO which includes the SSO as a LS operator is proposed. In it, each solution is randomly updated either via HHO or SSO operators which is determined by a probability calculated from the fitness of the individual. The problem of scheduling jobs on machines with preventive maintenances is examined in \cite{A183_LEI2020} with the goal of minimising makespan. The authors propose a distributed ABC method for solving the problem, which divides the bees into several colonies that differ in the search strategies (neighbourhood search methods) that are used. The problem of minimising the total weighted tardiness with job release times and setup times is considered in \cite{A197_MARINHODIANA2020}. The authors perform an analysis of the inclusion of VND in different metaheuristics (IGS, ABC, GA) instead of a using a simple LS operator. The authors test three neighbourhood operators and three VND strategies, and examined the influence of different parameters like the order of the application of the neighbourhood operators. The general conclusion was that integration of VND lead to significantly better results when compared to the original algorithms.  

In \cite{A200_Orts2020} the authors consider a GA for the makespan minimisation. The proposed algorithm has been tested for scheduling on several clusters when solving a difficult problem from statistical mechanics (calculating an active microrheology model). The authors also apply two DRs for scheduling, however, they cannot match the performance of GAs. Optimising the total weighted tardiness with rework processes is examined in \cite{A214_WANG2020}. The authors propose that the problem, which is stochastic in a way that it is not known which tasks will have to be reworked, is transformed into several deterministic problems using task estimation. These models are then solved either by using metaheuristics (SA and GA), or a simple DR. In \cite{A216_YEPESBORRERO2021} the authors consider a problem with setup times and resource constraints. In this problem the resources are not tied with processing jobs, but rather with setup times. The goal is to minimise the makespan and maximum resource consumption. The authors apply several MO methods like NSGA-II, MO iterated greedy search (MOIGS), and restarted iterated Pareto algorithm (RIPG). Additionally, they also propose a truncated RIPG (T-RIPG) method which incorporates several improvements over the original. 

A study on preventive maintenances is provided in \cite{A217_Ghaleb2020}. The authors consider two problem types, fixed maintenances (the time when they occur is given), or distributing maintenances to either maximise the availability of machines or their reliability. The authors consider that machines deteriorate over time, and that at a certain point they can break down and a corrective maintenance needs to be performed. The objective is to minimise the total cost of production (due to maintenance costs and late deliveries). The authors apply SA to solve the considered problem. In \cite{A218_Khanh2021} the authors consider minimising the sum of the makespan and total earliness and tardiness criterion with setup times. The authors propose a simple DR that assigns the jobs closest to their due dates. This might not be optimal and as such the authors propose that a GA is used to generate the sequence of jobs and the proposed DR their assignment to machines. In \cite{A255_Wang2020} the authors consider a resource constrained bi-objective problem. The considered resources represent raw materials which are stored in special containers. When the containers are opened, the resource it contained starts to deteriorate until a certain deadline when it becomes unusable and another container needs to be opened. The authors propose a MO hybrid PSO with a greedy solution construction and embedded LS operators for the minimisation of the total completion time and material cost. The method is compared to other MO algorithms and shows a better performance. 

In \cite{A239_Pinheiro2020} deal with a resource constrained scheduling problem with setup times in which the total tardiness objective is minimised. In the considered problem the resources are being supplied to machines with a constant rate. The authors apply SA and IG to solve the problem. Both methods use a simple DR for constructing the initial solution. A problem from the cloud computing domain is examined in \cite{A241_Bhardwaj2020}. In it, each job consists out of several tasks which need to be executed in a sequential manner. Two independent optimisation objectives are considered, makespan and total completion time minimisation. The authors propose the application of a DR, and ACO with LS operators. The makespan minimisation problem is considered in \cite{A4_EWEES2021}. The authors apply a SSO which includes the FA. The proposed method works in a way that certain probability (based on the fitness value) is calculated and based on it either the operators from SSO or FA are used. 

In \cite{A191_Lin2021} the authors consider a scheduling problem in which additional burn-in operations are performed. The equipment that performs these operations is scarce and represents the bottleneck of the system. The authors consider release times, machine eligibility constraints, machine availabilities, and setup times. The objective is to optimise the makespan and violations for burn-in operations. The authors apply a population based SA, which uses VND as the LS operator. The problem of makespan minimisation subject to release times, setup times, and renewable resources is examined in \cite{A199_pr9040654}. The authors propose the application of a modified HS. In the method the solution encoding was adapted to consider the resource constraints. The modified HS also includes several new concepts in the algorithm compared to the original version. 

A modified WOA for the problem considering setup times and makespan minimisation is investigated in \cite{A215_Alqaness2021}. In his study the WOA is combined with the FA in a way that each individual (depending on its fitness) is either modified using the operators from WOA or FA. In \cite{A219_Nanthapodej2021} a weighted sum of the total energy cost, number of tardy jobs and makespan is minimised. The authors consider machine eligibility constraints and the maximum number of tardy jobs that are allowed in the schedule. A VNS method with adaptive search is applied, and several new neighbourhood strategies are proposed and tested. In \cite{A236_su2021} the authors focus on a similar problem. The optimisation objective is the minimisation of the weighted sum of makespan and total energy consumption. The problem includes the machine eligibility constraint and a constraint that the execution time between all machines are not greater than a given threshold, which serves to obtain a similar load across all machines. The authors apply DE coupled with VNS which is performed after its mutation phase. VNS uses several destructions moves to remove job allocations and then repair moves to construct feasible schedule once again.

In \cite{A238_JOVANOVIC2021} the authors focus on the scheduling problems which includes setup times and with makespan minimisation. The authors apply FSS which is based on GRASP. The main difference between these methods is that in FSS some elements are fixed and it is expected that the final solution contains such elements. In \cite{A247_CHENG2021} the authors focus on the same problem, but apply an unsupervised learning based ABC algorithm. This method applies k-means clustering to group jobs so that the setup times between the consecutively executed jobs is decreased. The authors also introduce a LS operator in the form of the IG algorithm. In \cite{A237_ZHANG2021} a problem with setup times, additional worker resources, and learning effects is considered. In this problem the workers perform setup tasks and become more skilled over time which means that they perform those tasks in a shorter time. Additionally, all workers are not equal, and some perform better than others. The objective is to minimise the makespan and total energy consumption (machines consume different amount of energy when being idle or processing jobs). The authors apply CEA to solve the problem. The method uses a three vector solution for determining the job sequence, machine allocation, and worker allocation. Additionally, the initial solution is generated using simple DRs. In \cite{A256_Pan2021} the authors consider the bi-criteria objective of minimising the total tardiness and total energy consumption. This problem considers that machines are divided into several factories, and can have different processing speeds by which they can execute jobs. However, similar as in \cite{A9_Lei2020}, the authors disregard the placement of machines in factories. The authors apply a two-population based algorithm which uses NSGA-II and DE in cooperation to evolve the solutions. The algorithms uses two heuristics to create initial population, and also integrates two LS methods in the search process.

\section{Classification of research}
\label{sec:class}

In this section all the research will be roughly classified into three groups to obtain a better notion of the research which has been performed in each area. The classification is carried out by grouping the research based on the optimised criteria, additional constraints that are considered, and the solution methods used for solving the problem. 

Table \ref{tbl:criteria} shows the classification of the research based on the optimised criteria. In around 60\% of studies the makespan criterion is optimised, which makes it the most investigated objective for the UPMSP. The dominance of this objective is not unexpected, as reducing the total duration of the schedule has always been of importance. The second most popular optimisation objective is the total (weighted) tardiness, which is optimised in around 26\% of the studies. Therefore, the second most popular objective is already more than twice times less investigated. However, this objective is important in cases when due dates are defined for problems, and in such cases this objective is almost always considered. The only other criteria that are considered in more than 10\% of studies are the total (weighted) flowtime with share of around 19\%, and total weighted earliness and tardiness with a share of around 10\%. The total weighted earliness and tardiness objective is important in problems where the executed jobs need to be delivered as close to their due dates as possible, like in cases where the jobs represent goods that can get spoiled. All other objectives were rarely considered, with some very specialised objectives being examined in only a single study. As can be seen, most studies (around 85\%) focus exclusively on optimising standard scheduling criteria. Two non-standard objectives which received more attention are production cost and total energy consumption. The production cost objective was quite popular because in several studies it was important to also minimise a certain cost incurred by using additional resources or similar. On the other hand total energy consumption became more popular in recent years to improve energy efficiency as green manufacturing has been gaining more importance in the industry. In the end, it is clear that the completion time based criteria gained the most attention until now. 

The table also includes entries denoted as WS and MO in which several objectives are optimised simultaneously. WS outlines the papers which optimise several criteria using a weighted sum of objectives and thus focusing on obtaining a single objective. However, such an approach usually has certain drawbacks, since the weights of each objective need to be fine-tuned to obtain a solution which provides a good balance between all optimised objectives. On the other hand, MO denotes papers which deal with optimisation of several objectives simultaneously using specific algorithms. These algorithms are adapted so that they do not obtain only a single solution, but rather a set of solutions which provide a different trade-off between the optimised objectives. Both problem types are investigated in around 10\% of research. In most cases two objectives were optimised simultaneously, although some studies considered even 3 or more objectives. Even though the share of studies which deal with MO optimisation is still quite small, around half of them were published during the last several years. This shows that MO problems are becoming more studied in the UPMSP and that such a trend could continue in the future as MO is gaining more importance and new advances are being made it in. 

\begin{table}[]
	\caption{Research classification based on the optimised criteria}
	\label{tbl:criteria}
	\begin{tabular}{@{}lp{0.85\columnwidth}@{}}
		\toprule
		Criterion & References \\ \midrule
	$C_{max}$	&   
 \cite{A165_Ibarra1977}, 
 \cite{A38_De1980},
 \cite{A80_HARIRI1991},
 \cite{A207_MAHESWARAN1999},
 \cite{A245_Braun1999}, 
 \cite{A18_BRAUN2001}, 
  \cite{A243_Wu2001}, 
 \cite{A61_Kim2004}, 
  \cite{A212_Xhafa2007}, 
 \cite{A248_LUO2007}, 
 \cite{A29_Munir2008},
  \cite{A19_TSENG2009}, 
   \cite{A54_Chen2009}, 
  \cite{A208_Izakian2009},  
  \cite{A97_Liu2011},
  \cite{A17_RAMEZANIAN2012},
 \cite{A30_Rafsanjani2012}, 
  \cite{A249_Briceo2012},
  \cite{A90_YangKuei2013},
\cite{A114_Wang2012},
   \cite{A166_LI2013}, 
    \cite{A70_Santos}, 
     \cite{A40_ARROYO2017}, 
 \cite{A1_DURASEVIC2018}, 
 \cite{A213_Zarook2021},
  \cite{A225_DURASEVIC2016}, 
   \cite{A231_Durasevi2017},
  \cite{A228_Durasevi2017}, 
   \cite{A233_Durasevi2019}, 
  \cite{A251_SURESH1996}, 
   \cite{A65_HERRMANN1997},
    \cite{A57_DHAENENSFLIPO2001}, 
     \cite{A195_Salem2002},
       \cite{A49_Chen2004}, 
      \cite{A163_DOLGUI2009}, 
  \cite{A108_LIN2011},
   \cite{A124_Xu2015}, 
 \cite{A75_FanjulPeyro2017},
  \cite{A125_VILLA2018}, 
  \cite{A2_GLASS1994},  
 \cite{A13_PIERSMA199611}, 
 \cite{A104_SURESH1996},
 \cite{A102_Srivastava1998}, 
  \cite{A22_Anagnostopoulos2002},
 \cite{A50_COCHRAN20031087}, 
 \cite{A88_PENG2004}, 
  \cite{A246_Ritchie2003},
 \cite{A87_Gao2005}, 
 \cite{A25_Helal2006}, 
 \cite{A157_Rabadi2006},
 \cite{A188_GUO2007}, 
 \cite{A59_dePaula2007}, 
 \cite{A86_Xu2007}, 
 \cite{A137_Ravetti2007}, 
 \cite{A94_Arnaout2008}, 
 \cite{A20_Arnaout2009}, 
 \cite{A41_Arnaout2012}, 
 \cite{A77_VALLADA2011612}, 
 \cite{A78_Gao2008}, 
  \cite{A16_Charalambous}, 
 \cite{A72_FANJULPEYRO201055}, 
 \cite{A187_GAO2010}, 
 \cite{A39_BALIN20116814}, 
 \cite{A51_CHANG2011}, 
 \cite{A73_FANJULPEYRO2011301}, 
 \cite{A74_FANJULPEYRO2012}, 
 \cite{A106_VALLADA2011},  
 \cite{A11_Lin2013}, 
  \cite{A254_Niu2011}, 
 \cite{A71_Fleszar2011},  
 \cite{A82_HADDAD}, 
 \cite{A85_Ying2010}, 
 \cite{A168_KESKINTURK2012}, 
 \cite{A66_CAPPADONNA2013}, 
 \cite{A202_Capadonna2012}, 
 \cite{A68_Costa2013}, 
 \cite{A111_LIN2013}, 
 \cite{A115_WANG2013}, 
 \cite{A142_Low2013}, 
 \cite{A145_TORABI2013}, 
 \cite{A172_BANDYOPADHYAY2013}, 
 \cite{A206_Diana}, 
  \cite{A69_DIANA201594},
 \cite{A27_Cota}, 
 \cite{A42_Celano2008}, 
 \cite{A46_AvalosRosales2014}, 
 \cite{A47_Caniyilmaz2014}, 
 \cite{A181_Kayvanfar}, 
 \cite{A112_LIN2014},
 \cite{A123_Eroglu2014}, 
 \cite{A127_SELS2015}, 
 \cite{A167_LIAO2014}, 
 \cite{A173_RAMBOD2014}, 
  \cite{A222_Haddad2014}, 
 \cite{A189_Jabbar2014}, 
 \cite{A113_Lin2015}, 
  \cite{A136_Lin2016},  
 \cite{A132_Ebrahimi2015}, 
 \cite{A133_Liao2016}, 
 \cite{A179_LIN2015}, 
 \cite{A24_AFZALIRAD2016}, 
 \cite{A33_Wang}, 
 \cite{A35_ARROYO2017}, 
 \cite{A62_OZPEYNIRCI2016}, 
 \cite{A119_ZHENG2016}, 
  \cite{A120_Zheng2018},	
 \cite{A139_Santos2016},
 \cite{A141_JOO2017},
 \cite{A176_Low2016}, 
 \cite{A230_Durasevic2016}, 
 \cite{A96_Tozzo2018}, 
 \cite{A153_SHAHIDIZADEH2017}, 
 \cite{A194_SHAHVARI2017}, 
 \cite{A210_MANUPATI2017}, 
 \cite{A221_Cota2017}, 
 \cite{A184_LU2018}, 
  \cite{A253_DIANA2017},
  \cite{A190_Ezugwu2018}, 
 \cite{A31_EZUGWU2019}, 
 \cite{A26_AVALOSROSALES2018364},
 \cite{A28_ZHOU2018}, 
 \cite{A32_Afzalirad2015}, 
 \cite{A143_WU2019}, 
 \cite{A12_Ezugwu2018}, 
 \cite{A3_VALLADA2019}, 
 \cite{A9_Lei2020}, 
 \cite{A10_breu2020}, 
 \cite{A14_Wang2019}, 
 \cite{A15_ALHARKAM}, 
 \cite{A223_Jouhari2019}, 
 \cite{A234_YEPESBORRERO2020112959}, 
 \cite{A6_TERZI2020},
 \cite{A8_Lei2020},
 \cite{A34_Arnaout2019},
  \cite{A224_VLASIC2020},  
 \cite{A91_Jouhari2020}, 
 \cite{A183_LEI2020}, 
 \cite{A200_Orts2020}, 
 \cite{A216_YEPESBORRERO2021}, 
 \cite{A218_Khanh2021}, 
 \cite{A241_Bhardwaj2020}, 
 \cite{A4_EWEES2021}, 
 \cite{A191_Lin2021}, 
 \cite{A199_pr9040654}, 
 \cite{A215_Alqaness2021}, 
 \cite{A219_Nanthapodej2021}, 
 \cite{A236_su2021}, 
 \cite{A238_JOVANOVIC2021}, 
 \cite{A247_CHENG2021}, 
 \cite{A237_ZHANG2021}\\
	$Etwt$ & 
	 \cite{A1_DURASEVIC2018},
	  \cite{A231_Durasevi2017}, 
	\cite{A36_Bank2001},  
	 \cite{A211_POLYAKOVSKIY2014}, 
 \cite{A180_JOU2005}, 
 \cite{A89_MIN2006},
 \cite{A156_Raja2008}, 
 \cite{A78_Gao2008}, 
 \cite{A187_GAO2010},  
 \cite{A181_Kayvanfar}, 
 \cite{A100_2015}, 
 \cite{A201_DECMNOGUEIRA2014}, 
 \cite{A64_Zeidi2017}, 
 \cite{A153_SHAHIDIZADEH2017}, 
 \cite{A203_Abedi2017}, 
 \cite{A205_CHENG2017}, 
 \cite{A244_Kramer2017}, 
 \cite{A7_Ark2019}, 
 \cite{A218_Khanh2021}\\ 
	$Ft$ & 		
  \cite{A129_Randhawa1995}, 
 \cite{A212_Xhafa2007}, 
 \cite{A208_Izakian2009},
    \cite{A126_Ruiz2011},  
\cite{A30_Rafsanjani2012}, 
 \cite{A148_Strohhecker2016}, 
 \cite{A1_DURASEVIC2018}, 
  \cite{A225_DURASEVIC2016}, 
\cite{A231_Durasevi2017}, 
\cite{A228_Durasevi2017}, 
 \cite{A233_Durasevi2019}, 
 \cite{A99_Randhawa1997},
 \cite{A182_LEE2014}, 
 \cite{A198_Yildrim2007},
 \cite{A147_TAVAKKOLIMOGHADDAM2009}, 
 \cite{A150_Siepak2014}, 
 \cite{A140_Joo2015}, 
 \cite{A186_SALEHIMIR2016}, 
 \cite{A230_Durasevic2016}, 
 \cite{A64_Zeidi2017}, 
 \cite{A210_MANUPATI2017},
  \cite{A193_Arroyo2019}, 
   \cite{A224_VLASIC2020},
   \cite{A255_Wang2020},
 \cite{A241_Bhardwaj2020} 
            \\ 
	$Fwt$ & 
	 \cite{A93_WENG2001},
 \cite{A192_Arnaout2005},
 \cite{A138_Arnaout2006},
 \cite{A90_YangKuei2013},
  \cite{A1_DURASEVIC2018},
   \cite{A231_Durasevi2017}, 
  \cite{A108_LIN2011}, 
 \cite{A50_COCHRAN20031087}, 
 \cite{A170_Chavez}, 
 \cite{A58_Chyu2010}, 
 \cite{A67_Chyu2009}, 
 \cite{A131_Mehravaran2011}, 
 \cite{A45_BOZORGIRAD2012}, 
  \cite{A155_Rodriguez2012}, 
 \cite{A177_Rodriguez2012}, 
 \cite{A48_Bitar2014}, 
 \cite{A111_LIN2013}, 
 \cite{A145_TORABI2013}, 
 \cite{A169_Chang2014}, 
 \cite{A178_RODRIGUEZ2013}, 
 \cite{A204_Chen2013}, 
 \cite{A113_Lin2015}, 
 \cite{A136_Lin2016}, 
 \cite{A152_Shahvari2015}, 
 \cite{A161_AFZALIRAD2017}, 
 \cite{A151_Shahvari2017}, 
 \cite{A244_Kramer2017},
  \cite{A117_WANG2019}  \\
	$F_{max}$ &
	 \cite{A1_DURASEVIC2018},
	\cite{A231_Durasevi2017}
 \\
	$ML$		&    
	\cite{A129_Randhawa1995}, 
	 \cite{A212_Xhafa2007}, 
 \cite{A1_DURASEVIC2018},
 \cite{A231_Durasevi2017}, 
 \cite{A88_PENG2004}, 
 \cite{A180_JOU2005}, 
 \cite{A210_MANUPATI2017} 
          \\ 
	$TT$	&      
	\cite{A103_SURESH1994}, 
	\cite{A99_Randhawa1997},
	 \cite{A158_PEREZGONZALEZ2019}, 
 \cite{A84_KIM2002223}, 
  \cite{A88_PENG2004}, 
 \cite{A21_CHEN2006},
 \cite{A164_Kim2006}, 
 \cite{A137_Ravetti2007},
 \cite{A171_Kim2007}, 
 \cite{A52_Chen2008}, 
 \cite{A107_Lin2010}, 
 \cite{A196_akar2012}, 
 \cite{A135_Liu2013},
 \cite{A175_Lee2013}, 
 \cite{A42_Celano2008}, 
 \cite{A47_Caniyilmaz2014}, 
 \cite{A122_Zeidi2015},
  \cite{A210_MANUPATI2017}, 
 \cite{A220_Pan2018}, 
 \cite{A14_Wang2019}, 
 \cite{A8_Lei2020}, 
 \cite{A239_Pinheiro2020},
 \cite{A256_Pan2021}     \\
	$TWT$ & 
  \cite{A130_Na2006}, 
   \cite{A121_Zhang2006}, 
    \cite{A19_TSENG2009},
    \cite{A90_YangKuei2013},
     \cite{A110_Kuei2014},
      \cite{A1_DURASEVIC2018},  
      \cite{A225_DURASEVIC2016}, 
       \cite{A231_Durasevi2017}, 
     \cite{A228_Durasevi2017}, 
      \cite{A233_Durasevi2019}, 
      \cite{A229_DURASEVIC2020},  
     \cite{A227_Durasevi2020}, 
     \cite{A226_JAKLINOVIC2021}, 
 \cite{A98_KIM2003173},
  \cite{A108_LIN2011}, 
 	\cite{A50_COCHRAN20031087}, 
 	 \cite{A252_LOGENDRAN2004}, 
 \cite{A79_CAO2005}, 
 \cite{A59_dePaula2007},
 \cite{A118_Zhou2007}, 
 \cite{A83_Klemmt},
 \cite{A58_Chyu2010}, 
 \cite{A67_Chyu2009}, 
 \cite{A131_Mehravaran2011}, 
 \cite{A43_Bilyk2010}, 
 \cite{A45_BOZORGIRAD2012}, 
 \cite{A109_Lin2013}, 
 \cite{A111_LIN2013}, 
 \cite{A145_TORABI2013}, 
 \cite{A172_BANDYOPADHYAY2013}, 
 \cite{A95_Liang2015}, 
 \cite{A113_Lin2015}, 
 \cite{A136_Lin2016}, 
 \cite{A152_Shahvari2015}, 
 \cite{A179_LIN2015}, 
 \cite{A161_AFZALIRAD2017}, 
 \cite{A230_Durasevic2016},
 \cite{A151_Shahvari2017}, 
 \cite{A60_DIANA2018}, 
 \cite{A44_BEKTUR2019}, 
 \cite{A232_VLASIC2019}, 
 \cite{A235_SOLEIMANI2020}, 
 \cite{A244_Kramer2017}, 
 \cite{A224_VLASIC2020}, 
 \cite{A197_MARINHODIANA2020}, 
 \cite{A214_WANG2020}\\
	$T_{max}$ & 
	\cite{A1_DURASEVIC2018},
	 \cite{A231_Durasevi2017}, 
	\cite{A104_SURESH1996}, 
 \cite{A53_Chen2006}
 \\
	$Ut$ & 		
	\cite{A129_Randhawa1995}, 
	 \cite{A250_GolcondaDO04},
	\cite{A114_Wang2012},
	\cite{A99_Randhawa1997},
 \cite{A128_SILVA2006}, 
 \cite{A147_TAVAKKOLIMOGHADDAM2009}            \\ 
	$Uwt$ & 
	 \cite{A54_Chen2009}, 
	  \cite{A1_DURASEVIC2018}, 
	  \cite{A225_DURASEVIC2016}, 
	 \cite{A231_Durasevi2017}, 
	 \cite{A228_Durasevi2017}, 
	 \cite{A233_Durasevi2019},
	\cite{A159_Rojanasoonthon2005}, 
 \cite{A56_Chen2008}, 
 \cite{A55_Chen2011}, 
 \cite{A230_Durasevic2016}, 
 \cite{A224_VLASIC2020}
 \cite{A219_Nanthapodej2021}, 
 \\
	$COST$	&  
 \cite{A19_TSENG2009}, 
  \cite{A185_Li2015}, 
	\cite{A57_DHAENENSFLIPO2001},
	\cite{A79_CAO2005}, 
 \cite{A172_BANDYOPADHYAY2013},
 \cite{A153_SHAHIDIZADEH2017},
 \cite{A194_SHAHVARI2017},
 \cite{A203_Abedi2017}, 
 \cite{A240_LU2017}, 
 \cite{A217_Ghaleb2020},
 \cite{A255_Wang2020}, 
 \cite{A191_Lin2021}             \\ 
	$N_{jit}$ & \cite{A209_Jolai2009}\\ 
	$R_u$ & \cite{A126_Ruiz2011}, \cite{A216_YEPESBORRERO2021}\\
	$Tl$ & \cite{A160_Afzalirad2016}\\
	$TEC$ & 
 \cite{A162_CHE2017}, 
  \cite{A95_Liang2015}, 
	\cite{A120_Zheng2018}, 
 \cite{A143_WU2019}, 
 \cite{A220_Pan2018}, 
 \cite{A235_SOLEIMANI2020}, 
 \cite{A219_Nanthapodej2021}, 
 \cite{A236_su2021}, 
 \cite{A237_ZHANG2021},
 \cite{A256_Pan2021}\\
	$Ts$ & \cite{A148_Strohhecker2016} \\
		WS			&  
\cite{A54_Chen2009},
\cite{A126_Ruiz2011},
\cite{A114_Wang2012},
\cite{A57_DHAENENSFLIPO2001},
\cite{A180_JOU2005},
\cite{A79_CAO2005},
\cite{A59_dePaula2007},
\cite{A137_Ravetti2007},
\cite{A78_Gao2008},
\cite{A187_GAO2010},
\cite{A131_Mehravaran2011},
\cite{A45_BOZORGIRAD2012},
\cite{A47_Caniyilmaz2014},
\cite{A181_Kayvanfar},
\cite{A95_Liang2015},
\cite{A152_Shahvari2015},
\cite{A151_Shahvari2017},
\cite{A153_SHAHIDIZADEH2017},
\cite{A203_Abedi2017},
\cite{A235_SOLEIMANI2020},
\cite{A218_Khanh2021},
\cite{A191_Lin2021},
\cite{A219_Nanthapodej2021},
\cite{A236_su2021}      \\ 
		MO			&  
\cite{A231_Durasevi2017},
\cite{A99_Randhawa1997},
\cite{A104_SURESH1996},
\cite{A50_COCHRAN20031087},
\cite{A147_TAVAKKOLIMOGHADDAM2009},
\cite{A58_Chyu2010},
\cite{A67_Chyu2009},
\cite{A111_LIN2013},
\cite{A145_TORABI2013},
\cite{A172_BANDYOPADHYAY2013},
\cite{A113_Lin2015},
\cite{A136_Lin2016},
\cite{A179_LIN2015},
\cite{A120_Zheng2018},
\cite{A161_AFZALIRAD2017},
\cite{A64_Zeidi2017},
\cite{A96_Tozzo2018},
\cite{A194_SHAHVARI2017},
\cite{A210_MANUPATI2017},
\cite{A143_WU2019},
\cite{A220_Pan2018},
\cite{A14_Wang2019},
\cite{A8_Lei2020},
\cite{A216_YEPESBORRERO2021},
\cite{A255_Wang2020},
\cite{A237_ZHANG2021},
\cite{A256_Pan2021}       \\ 
		\bottomrule
	\end{tabular}
\end{table}

Table \ref{tbl:properties} shows the additional constraints that have been considered in the reviewed studies. Most research has clearly focused on problems which include setup times, with around 50\% of the research focusing on such problems. Since setup times usually have a direct effect on the performance of the scheduling system \cite{ALLAHVERDI2016}, it is good that most research in the UPMSP does take them into account. The second most commonly considered constraint are job release times with a share of around 23\%. Release times are especially common in cases when dynamic scheduling problems are considered, in which it is not known when jobs are released into the system and as such it is not possible to create the schedule in advance. Some other more frequently considered objectives are machine eligibility, additional resources, batch scheduling, which are considered in between 10\% to 14\% of the studies. Other constraints are only sparsely considered, although this does not mean that they are less important. For example machine breakdowns and precedence constraints were not considered that often, but such constraints can appear in real world problems. Naturally, it should be outlined that the problem types are more numerous than denoted with this classification, since some constraints have different types. For example, the setup times can depend on various things (previous jobs, next job, machine, assigned worker), or resource constraints can be of different types (renewable or nonrenewable), characteristics, or quantities. Therefore the number of different problem variants is even larger. 

It is interesting to observe that around 23\% of research focused on the most basic problem without considering any additional properties. However, a great deal of such research was done in the earlier research performed in the UPMSP and recent studies almost always include at least on additional property. Not only that, but an increasing number of studies deal with problems which include several additional constraints. This is especially the case in studies which model a real scheduling problem as an UPMSP, as in those cases several additional constraints have to be taken into account while scheduling. Therefore, it is expected that in the future studies will focus even more on optimising problems with several different constraints, and that even more new constraint variants will be defined.

\begin{table}[]
		\caption{Research classification based on additional properties}
	\label{tbl:properties}
	\begin{tabular}{@{}lp{0.8\columnwidth}@{}}
		\toprule
		Constraints & References \\ \midrule
None	&   
\cite{A165_Ibarra1977}, 
 \cite{A38_De1980}, 
 \cite{A80_HARIRI1991}, 
 \cite{A207_MAHESWARAN1999}, 
 \cite{A245_Braun1999}, 
 \cite{A18_BRAUN2001}, 
  \cite{A243_Wu2001}, 
    \cite{A250_GolcondaDO04}, 
   \cite{A212_Xhafa2007}, 
  \cite{A248_LUO2007}, 
  \cite{A29_Munir2008},   
  \cite{A208_Izakian2009}, 
  \cite{A30_Rafsanjani2012},
   \cite{A249_Briceo2012}, 
   \cite{A70_Santos}, 
 \cite{A103_SURESH1994}, 
   \cite{A108_LIN2011}, 
   \cite{A211_POLYAKOVSKIY2014}, 
     \cite{A162_CHE2017}, 
   \cite{A2_GLASS1994}, 
  \cite{A13_PIERSMA199611}, 
  \cite{A104_SURESH1996},
 \cite{A102_Srivastava1998}, 
 \cite{A50_COCHRAN20031087},
 \cite{A88_PENG2004}, 
 \cite{A246_Ritchie2003}, 
 \cite{A79_CAO2005}, 
 \cite{A87_Gao2005},
 \cite{A188_GUO2007},
 \cite{A118_Zhou2007},
 \cite{A170_Chavez}, 
    \cite{A16_Charalambous}, 
 \cite{A72_FANJULPEYRO201055}, 
 \cite{A39_BALIN20116814}, 
 \cite{A73_FANJULPEYRO2011301}, 
 \cite{A74_FANJULPEYRO2012}, 
 \cite{A209_Jolai2009}, 
 \cite{A155_Rodriguez2012}, 
 \cite{A177_Rodriguez2012}, 
 \cite{A109_Lin2013}, 
 \cite{A111_LIN2013}, 
 \cite{A169_Chang2014}, 
 \cite{A178_RODRIGUEZ2013},
 \cite{A127_SELS2015}, 
 \cite{A150_Siepak2014}, 
 \cite{A189_Jabbar2014}, 
 \cite{A113_Lin2015}, 
 \cite{A136_Lin2016}, 
 \cite{A117_WANG2019}, 
 \cite{A220_Pan2018}, 
 \cite{A9_Lei2020}, 
 \cite{A244_Kramer2017},
 \cite{A8_Lei2020}, 
 \cite{A91_Jouhari2020}, 
 \cite{A200_Orts2020}        \\
$d_j=D$ & 
\cite{A36_Bank2001},
 \cite{A89_MIN2006}, 
 \cite{A7_Ark2019} \\
$\bar{d}_j=D$ & \cite{A182_LEE2014} \\
$s$		&     
\cite{A129_Randhawa1995}, 
 \cite{A93_WENG2001}, 
 \cite{A192_Arnaout2005}, 
\cite{A138_Arnaout2006}, 
 \cite{A130_Na2006},
  \cite{A121_Zhang2006}, 
  \cite{A19_TSENG2009},
 \cite{A54_Chen2009},
  \cite{A126_Ruiz2011},
  \cite{A114_Wang2012},
   \cite{A110_Kuei2014},   
   \cite{A226_JAKLINOVIC2021}, 
 \cite{A99_Randhawa1997}, 
 \cite{A57_DHAENENSFLIPO2001}, 
  \cite{A98_KIM2003173}, 
 \cite{A49_Chen2004}, 
  \cite{A128_SILVA2006}, 
   \cite{A163_DOLGUI2009}, 
    \cite{A182_LEE2014}, 
     \cite{A148_Strohhecker2016}, 
    \cite{A158_PEREZGONZALEZ2019},
 \cite{A84_KIM2002223},
 \cite{A22_Anagnostopoulos2002}, 
 \cite{A180_JOU2005},
 \cite{A159_Rojanasoonthon2005}, 
 \cite{A21_CHEN2006}, 
  \cite{A53_Chen2006}, 
 \cite{A25_Helal2006}, 
 \cite{A157_Rabadi2006}, 
 \cite{A164_Kim2006}, 
 \cite{A59_dePaula2007}, 
 \cite{A137_Ravetti2007}, 
 \cite{A171_Kim2007}, 
 \cite{A198_Yildrim2007}, 
 \cite{A56_Chen2008}, 
 \cite{A94_Arnaout2008}, 
 \cite{A20_Arnaout2009}, 
 \cite{A41_Arnaout2012}, 
 \cite{A156_Raja2008},  
 \cite{A52_Chen2008}, 
 \cite{A77_VALLADA2011612}, 
 \cite{A147_TAVAKKOLIMOGHADDAM2009}, 
 \cite{A58_Chyu2010}, 
 \cite{A67_Chyu2009}, 
 \cite{A51_CHANG2011}, 
 \cite{A106_VALLADA2011}, 
 \cite{A107_Lin2010}, 
 \cite{A131_Mehravaran2011}, 
 \cite{A45_BOZORGIRAD2012}, 
 \cite{A254_Niu2011}, 
 \cite{A55_Chen2011},
 \cite{A71_Fleszar2011},  
 \cite{A82_HADDAD}, 
 \cite{A85_Ying2010}, 
 \cite{A168_KESKINTURK2012}, 
 \cite{A66_CAPPADONNA2013}, 
 \cite{A202_Capadonna2012},
 \cite{A68_Costa2013},   
 \cite{A48_Bitar2014}, 
 \cite{A145_TORABI2013}, 
 \cite{A172_BANDYOPADHYAY2013}, 
 \cite{A175_Lee2013}, 
 \cite{A204_Chen2013}, 
 \cite{A206_Diana}, 
  \cite{A69_DIANA201594}, 
 \cite{A27_Cota}, 
 \cite{A42_Celano2008}, 
 \cite{A46_AvalosRosales2014},  
 \cite{A47_Caniyilmaz2014}, 
 \cite{A181_Kayvanfar}, 
 \cite{A100_2015}, 
 \cite{A112_LIN2014}, 
 \cite{A123_Eroglu2014}, 
 \cite{A167_LIAO2014}, 
 \cite{A173_RAMBOD2014}, 
 \cite{A201_DECMNOGUEIRA2014}, 
 \cite{A222_Haddad2014}, 
 \cite{A122_Zeidi2015}, 
 \cite{A133_Liao2016}, 
 \cite{A140_Joo2015}, 
 \cite{A152_Shahvari2015}, 
 \cite{A160_Afzalirad2016}, 
 \cite{A186_SALEHIMIR2016}, 
 \cite{A24_AFZALIRAD2016}, 
 \cite{A33_Wang}, 
 \cite{A139_Santos2016}, 
 \cite{A141_JOO2017}, 
 \cite{A161_AFZALIRAD2017},
  \cite{A64_Zeidi2017}, 
 \cite{A96_Tozzo2018}, 
 \cite{A151_Shahvari2017}, 
 \cite{A194_SHAHVARI2017}, 
 \cite{A210_MANUPATI2017}, 
 \cite{A221_Cota2017}, 
  \cite{A253_DIANA2017}, 
 \cite{A190_Ezugwu2018},
 \cite{A31_EZUGWU2019}, 
 \cite{A26_AVALOSROSALES2018364}, 
 \cite{A60_DIANA2018}, 
 \cite{A12_Ezugwu2018}, 
 \cite{A10_breu2020}, 
 \cite{A14_Wang2019},
 \cite{A15_ALHARKAM}, 
 \cite{A44_BEKTUR2019},  
 \cite{A223_Jouhari2019}, 
 \cite{A234_YEPESBORRERO2020112959}, 
 \cite{A235_SOLEIMANI2020}, 
 \cite{A244_Kramer2017}, 
 \cite{A6_TERZI2020}, 
 \cite{A34_Arnaout2019}, 
 \cite{A197_MARINHODIANA2020}, 
 \cite{A216_YEPESBORRERO2021}, 
 \cite{A218_Khanh2021},
 \cite{A4_EWEES2021}, 
 \cite{A191_Lin2021}, 
 \cite{A199_pr9040654},
 \cite{A215_Alqaness2021}, 
 \cite{A238_JOVANOVIC2021}, 
 \cite{A247_CHENG2021}, 
 \cite{A237_ZHANG2021}    \\
$brkdwn$		&      
 \cite{A226_JAKLINOVIC2021},
\cite{A251_SURESH1996}, 
 \cite{A252_LOGENDRAN2004}, 
 \cite{A45_BOZORGIRAD2012}, 
 \cite{A152_Shahvari2015}, 
 \cite{A151_Shahvari2017}, 
 \cite{A194_SHAHVARI2017}, 
 \cite{A217_Ghaleb2020}, 
 \cite{A191_Lin2021}     
 \\ 
$M_j$ & 
 \cite{A121_Zhang2006}, 
  \cite{A54_Chen2009}, 
  \cite{A114_Wang2012},
\cite{A226_JAKLINOVIC2021},
\cite{A99_Randhawa1997}, 
 \cite{A195_Salem2002},
  \cite{A128_SILVA2006},
  \cite{A163_DOLGUI2009},
  \cite{A158_PEREZGONZALEZ2019}, 
  \cite{A252_LOGENDRAN2004},  
 \cite{A159_Rojanasoonthon2005}, 
 \cite{A21_CHEN2006}, 
 \cite{A53_Chen2006}, 
 \cite{A198_Yildrim2007}, 
 \cite{A78_Gao2008}, 
 \cite{A187_GAO2010}, 
 \cite{A45_BOZORGIRAD2012}, 
 \cite{A115_WANG2013}, 
 \cite{A142_Low2013}, 
 \cite{A47_Caniyilmaz2014}, 
 \cite{A181_Kayvanfar},
 \cite{A173_RAMBOD2014}, 
 \cite{A140_Joo2015}, 
 \cite{A152_Shahvari2015}, 
 \cite{A160_Afzalirad2016}, 
 \cite{A24_AFZALIRAD2016}, 
 \cite{A161_AFZALIRAD2017}, 
 \cite{A176_Low2016},
  \cite{A64_Zeidi2017},
   \cite{A151_Shahvari2017},  
 \cite{A194_SHAHVARI2017}, 
 \cite{A203_Abedi2017}, 
 \cite{A44_BEKTUR2019}, 
 \cite{A191_Lin2021}, 
 \cite{A219_Nanthapodej2021}, 
 \cite{A236_su2021}
 \\
$M_{ded}$ & \cite{A182_LEE2014}, 
 \cite{A205_CHENG2017} \\
$r_j$ &  
 \cite{A61_Kim2004}, 
  \cite{A54_Chen2009}, 
   \cite{A17_RAMEZANIAN2012},
   \cite{A90_YangKuei2013},
   \cite{A114_Wang2012},
    \cite{A110_Kuei2014}, 
    \cite{A185_Li2015},
     \cite{A40_ARROYO2017}, 
    \cite{A1_DURASEVIC2018}, 
    \cite{A213_Zarook2021},
     \cite{A225_DURASEVIC2016}, 
     \cite{A231_Durasevi2017}, 
    \cite{A228_Durasevi2017},
     \cite{A233_Durasevi2019},
     \cite{A229_DURASEVIC2020}, 
    \cite{A227_Durasevi2020}, 
   \cite{A226_JAKLINOVIC2021}, 
\cite{A36_Bank2001},
 \cite{A252_LOGENDRAN2004},  
 \cite{A159_Rojanasoonthon2005}, 
 \cite{A164_Kim2006}, 
 \cite{A171_Kim2007}, 
 \cite{A83_Klemmt}, 
 \cite{A147_TAVAKKOLIMOGHADDAM2009}, 
 \cite{A11_Lin2013}, 
 \cite{A43_Bilyk2010}, 
 \cite{A45_BOZORGIRAD2012}, 
 \cite{A55_Chen2011}, 
 \cite{A196_akar2012}, 
 \cite{A145_TORABI2013}, 
 \cite{A167_LIAO2014}, 
 \cite{A173_RAMBOD2014}, 
 \cite{A95_Liang2015}, 
 \cite{A152_Shahvari2015},
 \cite{A160_Afzalirad2016},  
 \cite{A179_LIN2015},  
 \cite{A186_SALEHIMIR2016}, 
 \cite{A24_AFZALIRAD2016}, 
 \cite{A35_ARROYO2017}, 
 \cite{A161_AFZALIRAD2017}, 
 \cite{A230_Durasevic2016}, 
 \cite{A64_Zeidi2017}, 
 \cite{A151_Shahvari2017}, 
 \cite{A153_SHAHIDIZADEH2017}, 
 \cite{A194_SHAHVARI2017}, 
 \cite{A210_MANUPATI2017},  
 \cite{A15_ALHARKAM}, 
 \cite{A193_Arroyo2019}, 
 \cite{A232_VLASIC2019}, 
 \cite{A244_Kramer2017}, 
  \cite{A224_VLASIC2020},
 \cite{A197_MARINHODIANA2020},
  \cite{A191_Lin2021},
 \cite{A199_pr9040654}
 \\
$batch$ & 
 \cite{A192_Arnaout2005}, 
\cite{A138_Arnaout2006},
 \cite{A130_Na2006}, 
  \cite{A166_LI2013}, 
   \cite{A40_ARROYO2017}, 
   \cite{A213_Zarook2021},
 \cite{A98_KIM2003173}, 
 \cite{A128_SILVA2006}, 
  \cite{A163_DOLGUI2009},
   \cite{A124_Xu2015},  
 \cite{A84_KIM2002223}, 
  \cite{A252_LOGENDRAN2004}, 
 \cite{A86_Xu2007}, 
 \cite{A83_Klemmt}, 
 \cite{A45_BOZORGIRAD2012}, 
 \cite{A115_WANG2013}, 
 \cite{A42_Celano2008}, 
 \cite{A152_Shahvari2015}, 
 \cite{A35_ARROYO2017}, 
 \cite{A151_Shahvari2017}, 
 \cite{A153_SHAHIDIZADEH2017}, 
 \cite{A194_SHAHVARI2017}, 
 \cite{A184_LU2018}, 
 \cite{A240_LU2017}, 
 \cite{A28_ZHOU2018}, 
 \cite{A193_Arroyo2019} \\
$R$ & 
\cite{A126_Ruiz2011},
\cite{A49_Chen2004}, 
 \cite{A124_Xu2015}, 
 \cite{A75_FanjulPeyro2017}, 
  \cite{A125_VILLA2018}, 
 \cite{A21_CHEN2006}, 
 \cite{A53_Chen2006}, 
 \cite{A66_CAPPADONNA2013}, 
 \cite{A202_Capadonna2012}, 
 \cite{A68_Costa2013},
 \cite{A48_Bitar2014}, 
 \cite{A142_Low2013}, 
 \cite{A145_TORABI2013}, 
 \cite{A42_Celano2008}, 
 \cite{A24_AFZALIRAD2016}, 
 \cite{A62_OZPEYNIRCI2016}, 
 \cite{A119_ZHENG2016}, 
 \cite{A120_Zheng2018}, 
  \cite{A176_Low2016},
 \cite{A194_SHAHVARI2017}, 
 \cite{A210_MANUPATI2017}, 
 \cite{A32_Afzalirad2015}, 
 \cite{A3_VALLADA2019}, 
 \cite{A15_ALHARKAM}, 
 \cite{A234_YEPESBORRERO2020112959}, 
 \cite{A216_YEPESBORRERO2021}, 
 \cite{A255_Wang2020}
 \cite{A239_Pinheiro2020}, 
  \cite{A191_Lin2021},
 \cite{A199_pr9040654}, 
 \cite{A237_ZHANG2021}\\
$d$ & \cite{A159_Rojanasoonthon2005}, 
 \cite{A52_Chen2008}, 
 \cite{A204_Chen2013} \\
$p_c$ & \cite{A142_Low2013}, 
 \cite{A186_SALEHIMIR2016}, 
  \cite{A176_Low2016},
 \cite{A184_LU2018}, 
 \cite{A235_SOLEIMANI2020} \\
$j_f$ & \cite{A54_Chen2009}, 
\cite{A83_Klemmt}\\
$prec$ & 
 \cite{A97_Liu2011}, 
 \cite{A226_JAKLINOVIC2021},
 \cite{A65_HERRMANN1997},
\cite{A147_TAVAKKOLIMOGHADDAM2009}, 
 \cite{A196_akar2012}, 
 \cite{A135_Liu2013}, 
 \cite{A160_Afzalirad2016}, 
 \cite{A24_AFZALIRAD2016}, 
 \cite{A161_AFZALIRAD2017}, 
 \cite{A241_Bhardwaj2020} \\
$rwrk$ & \cite{A17_RAMEZANIAN2012},
\cite{A173_RAMBOD2014}, 
 \cite{A214_WANG2020} \\
$L$ & \cite{A43_Bilyk2010} \\
$M_d$ & \cite{A172_BANDYOPADHYAY2013}, 
 \cite{A203_Abedi2017}, 
 \cite{A217_Ghaleb2020}\\
$M_m$ & \cite{A132_Ebrahimi2015}, 
 \cite{A203_Abedi2017}, 
 \cite{A184_LU2018}, 
 \cite{A26_AVALOSROSALES2018364}, 
 \cite{A14_Wang2019}, 
 \cite{A183_LEI2020}, 
 \cite{A217_Ghaleb2020}  \\
$j_s$ & \cite{A40_ARROYO2017}, 
\cite{A213_Zarook2021},
 \cite{A35_ARROYO2017}, 
 \cite{A153_SHAHIDIZADEH2017}, 
 \cite{A194_SHAHVARI2017},
 \cite{A28_ZHOU2018},
 \cite{A193_Arroyo2019} \\
$Q_k$ & 
 \cite{A40_ARROYO2017}, 
 \cite{A213_Zarook2021},
  \cite{A35_ARROYO2017}, 
 \cite{A153_SHAHIDIZADEH2017},
  \cite{A194_SHAHVARI2017},
 \cite{A28_ZHOU2018},
 \cite{A193_Arroyo2019} \\
$M_s$ & \cite{A120_Zheng2018}, 
 \cite{A143_WU2019},
 \cite{A256_Pan2021} \\
\bottomrule
	\end{tabular}
\end{table}

Finally, Table \ref{tbl:methods} shows the application of different methods in the reviewed papers. Since many papers usually combine several methods, these studies are then classified in a way that they are enumerated aside each method that is applied. This is especially true for DRs which are often used  to generate initial solutions for metaheuristic methods. DRs are applied on average in every third paper, where in around half of these paper DRs are used in combination with other methods to improve their results. This is most often done by generating initial solution, or using DRs in some solution construction methods. This just provides further motivation to continue the research in designing better DRs, as they are often used either by themselves or in combination with other methods. In most cases, DRs were designed manually, however, there is a small number of studies in the last few years which propose that such methods are generated automatically, and such approach demonstrated promising results.  In comparison, other problem specific heuristics that are developed for the UPMSP make up only around 10\% of all research. This shows that for this type of problem the design of heuristic methods goes more into the direction of DRs due to their simplicity. 

Out of the metaheuristic methods, GAs are most commonly used, with around 33\% of research using GAs for solving this problem. Their popularity comes from the fact that GA is a quite powerful, yet simple and very flexible method which can be used for optimisation. Aside from GAs, it is evident that the next two most popular algorithms are usually simple single solution based metaheuristics like SA and TS, being used in around 15\% and 12\% of papers, respectively. This shows that even quite simple methods are commonly used and can obtain quite good results if designed well. LS based methods, like VNS, GRASP and similar are also very popular, either being used by themselves or in combination with other metaheuristic methods. Such methods have jointly been used in more than a quarter of all research. Naturally, simple LS operators by themselves have been used in even more research, because they are commonly included in different metaheuristic methods which by default do not use them. The reason as why those methods have been adopted so frequently is because of their simplicity and effectiveness on solving various problem types. 

Other methods received less attention, and were usually applied in only a few studies. Among these algorithms ACO was among the most frequently used. This is quite surprising as the algorithm needs some adaptation to be applied on the UPMSP. However, regardless of this, several studies applied this algorithm and focused especially on how to adapt the solution representation to the problem. Other classical metaheuristic methods which were used in several studies include ABC, DE, CLONALG, AIS, and PSO. However, each of them was used in less than 5\% of papers. Other methods which could not be classified to any of these categories are applied in around 15\% of the research. Most of these methods are different metaphor based metaheuristics that were proposed in the last several years. Therefore, most of the papers which use them have been published in recent years and focused on applying various new metaheuristics methods on the UPMSP, like FA, SCA, HHA, and similar. However, most of these methods are used only in a single study and have not achieved the same popularity and adaption as classical metaheuristic methods. However, it has yet to be seen if any of these methods will gain momentum, and gain a larger importance in the UPMSP. For now, it still seems that most research is still oriented towards using classical metaheuristic methods.

\scriptsize
\begin{table}[]
	\caption{Research classification based on the solution methods}
	\label{tbl:methods}
\begin{tabular}{@{}p{0.17\columnwidth}p{0.8\columnwidth}@{}}
		\toprule
		Algorithm & References \\
\bottomrule
	\end{tabular}
\end{table}
\normalsize

{
\makeatletter
\def\old@comma{,}
\catcode`\,=13
\def,{%
	\ifmmode%
	\old@comma\discretionary{}{}{}%
	\else%
	\old@comma%
	\fi%
}
\makeatother

A complete classification of the reviewer research including the problem variant, solution method, and short notes is given in Table \ref{tbl:class}.

\onecolumn
\scriptsize
	\begin{longtable}{@{}lp{0.3\textwidth}p{0.15\textwidth}p{0.45\textwidth}@{}}
		\caption{Classification of UPMSP research \label{tbl:class}}\\
		\toprule
		Reference                          & Scheduling problem                                                                                                                                                                                                                                                                                                                                                                                                                                                                                                                                                                                                                                                                                                                                                                                                                                                                                                   & Solution method                                                                      & Notes                                                                                                                                                                                                                         \\
\bottomrule
	\end{longtable}
\normalsize
}
\twocolumn

\section{Recent trends and future directions}
\label{sec:outlook}
\subsection{Problem definitions}
One important issue with the current research is the wide range of different problem instances that are used. This presents a quite large problem of the current research because it is difficult to compare the results across different studies. Quite often problem instances are generated, and although there are certain suggestions which are usually used to generate some parts of the problem instances, there are no guidelines or rules that would specify how to perform this in order to make it as general as possible. Naturally, a great problem here is the fact that there are many different problem types and variants that are considered. Several authors made their problem instances publicly available, which represents a good start in the direction of using a common problem set. However, these problem instances are usually designed for considering a general problem variant and therefore they cannot be use in all cases. Additionally, in most research smaller instances were used which include up to 200 job and 20 machines. However, larger instances should also be used to stress test the proposed methods and analyse their performance as the problem size increases. 

An important step in this direction would be to define a common set of problem instances which can be used in layers. This would mean that all the information for the different problem variants would be specified in them, however, the researchers could then use only the subset which is of interest to them. In this way, it would be possible to base all research on the same problems, and only use parts of those instances that are relevant. In addition, this would also allow other researchers to provide extensions to already existing datasets, and simply add additional constraints to an already existing basis. Important here would also be that such a problem set includes instances generated in different ways (to ensure that the methods are not overspecialised on a very narrow problem type), and include problem instances of different sizes to better test the generality and scalability of the proposed methods.

\subsection{Solution methods}
The review demonstrated that a wide range of solution methods were proposed, which range from quite simple heuristics, to complicated hybrid metaheuristics which combine several methods into one. All such algorithms offer different benefits and drawbacks, and it is up to the user to determine the one that should be used, which is not always trivial.

DRs have proven to be quite useful methods, especially in cases when a solution needs to be obtained in a small amount of time, or not all information about the problem is available. However, designing DRs of good quality requires good domain knowledge about the problem, and is usually a time consuming task. Luckily, at least for the standard criteria and problem types quite good DRs have been developed. However, as more exotic problem types are considered, it is more probable that an appropriate DR does not exits. Therefore, several studies have investigated how to automatically generate DRs for the different variants of the UPMSP. The ADDRs demonstrated a good performance, and usually outperformed MDDRs in many occasions. Therefore, further investigation in this area could lead to even better DRs that can be used for solving problems by themselves, or in combination with other methods. 

Many studies demonstrated that individual methods are usually not powerful enough to effectively solve the considered problem. Therefore, a lot of research used hybrid methods. This was mostly realised in two ways. One way was to apply DRs in initialisation of solutions, or in some algorithm operators when constructing the solution. The approach was to include different problem specific LS operators in metaheuristic methods. Both approaches have proven to be quite efficient. However, when combining such methods, several design choices need to be performed in the algorithms, for example which LS operators to use, when to apply them in the algorithm and similar. Some studies have already tried to provide answers to such questions, but still a lot of design choices which need to be performed. An interesting research direction would be to design methods that could by themselves learn which operators to apply in which situations, and relieve the designer of that choice. 

Recent years also saw the rise of new metaheuristic methods, which have also been applied for the UPMSP. Such new metaheuristics has often been criticised as they usually do not provide enough novelty or function quite similar as existing methods \cite{Sorensen2013}. In addition, there is also a line of research which considers hybrid metaheuristics which combine concepts of several methods into one. Although there is merit in such research directions, the application of hybrid or novel algorithms should be done sparingly and examined in detail to demonstrate their effectiveness compared to existing methods. In this case, a common dataset consisting out of several different problems is required to effectively test how effective these novel approaches are, since due to the no free lunch theorem it will always be possible to find a set of instances on which an algorithm will be more effective than others \cite{Wolpert1997}. Therefore, the goal of future research should not be to find the ideal metaheuristic for each specific UPMSP type, but rather to design methods which perform well over a wide range of problem types and gain better understanding which concepts are important in such algorithms. 

\subsection{Dynamic and stochastic problems}
Until now most problems were deterministic and static, which means that all the system parameters were known beforehand, and the schedule could be constructed up front and then simply executed at a later point in time. Unfortunately, in real world problems all information about the problem are not known up front, or unforeseen situations happen during the execution of the schedule (like machine breakdowns, or the arrival of new jobs) \cite{Ouelhadj2008}. As a result, the schedules obtained by metaheuristics can become invalid if unexpected situations like these happen. Therefore, it is required to design methods which build schedules incrementally and can this quickly react to changes in the system, or methods which when generate a schedule have the possibility of correcting it and in such a way to adapt the schedule for the unexpected situations. DRs are mostly easily applied to such a problem since they incrementally build the schedule. However, because of that the quality of the schedule they construct is limited. Therefore, one research direction is to improve the performance of DRs that they have a better overview on the problem and to reduce the gap between their results and those of metaheuristics. On the other hand, metaheuristic methods provide solutions of a better quality, however, it is more difficult to apply them for dynamic scheduling problems. Some studies tried to apply metaheuristics for problems which did include certain uncertainties (mostly concerned with rework processes), but but such research is still quite sparse. Therefore, an important research direction would be to put more focus on solving dynamic and stochastic scheduling problems, and adapt metaheuristics for unforeseen situations that can happen during the system execution. For example, one possibility would be to define correction methods which could be applied to correct results obtained by metaheuristics given an unexpected situation.

\subsection{Application for real world problems}
An important thing which needs to be considered is that the methods that are proposed can be applied for real world problems. Real world problems usually require that not only a single objective is optimised, but rather that several objectives are optimised simultaneously \cite{Heidi2008}. In that case a solution which represents a trade off between all objectives needs to be obtained. However, as was denoted in the literature review, only a handful of studies focused on MO problems, and a lot of studies used a weighted sum of several objectives. Therefore, it is imperative to put more focus on such problems. However, it is very likely that this direction will get a lot attention in the future since MO optimisation is gaining more momentum across different optimisation problems, and as such it is expected that research in the UPMSP will follow this trend. Additionally, most of the studies dealing with MO optimisation were published in recent years, which suggest that such a trend could continue further.

Apart from multiple objectives, another important thing is that real world problems usually are quite complicated and include many different and specific constraints. Until now most problems considered in the literature included only one or two additional constraints, and very often did not consider many situations which can happen in the real world. Most often the research focused on the setup time constraints which, although is very important, is probably the most easiest constraint to consider, as it requires virtually no adaptation in the solution methods or solution representation. On the other hand, certain scheduling properties received very little attention, which means that research on them is limited and there might be a lack of appropriate methods which consider such properties. This especially is true for some constraints which require more involvement in the design of the algorithm, like machine precedence constraints or machine breakdowns. Some recent papers, which modelled problems from the real world, focused on problems that included several additional constraints. Naturally such problems are not only more difficult to solve, and require that the solution methods is adapted to be able to consider such problems. 

\subsection{Green scheduling}  
Green manufacturing has become an increasingly important paradigm in recent years, which aims to reduce the impact of the industrial production on the environment \cite{Ahemad2013}. In that regard, scheduling also plays an important role as it directly influences the production environment \cite{AKBAR2018866}. Therefore, during the construction of the scheduled it is not only required to optimise the main objective, but also to focus on its impact on the environment. In the UPMSP the research until now mostly considered the effect on the environment through energy consumption that needed to be minimised. However, as green manufacturing is becoming more important, it is likely that more criteria and constraints will be introduced and will have to be considered during scheduling. Very likely the models that simulate the impact of such problems on the environment will become more realistic, and it will be required to consider more constraints than for classical scheduling problems. Therefore, this is one likely direction in which future research could be oriented.

\section{Conclusion}
\label{sec:conc}
This study provides a review of over 200 papers dealing with the application od heuristic and metaheuristic methods on the UPMSP. The review shows that this problem is tackled with a plethora of different methods, and that during the years several problem variants and optimisation criteria were considered. This has become especially evident in the last years, since around 70\% of research in this area has been published in the last 10 years, and around 40\% in the last 5 years. This shows that the interest for this problem is increasing, and the number of publications in the recent years show that such a trend could continue. As such, it is important to have a good overview of existing research to diminish the possibility of repeating research.

The goal of this review was twofold. The first goal was to provide an extensive overview of heuristic methods that are applied for solving the UPMSP. The focus was put equally both on problem specific heuristics, as well as metaheuristics, since in a lot of research these methods are used in synergy to increment each other. This overview presents the details of the different methods that were applied for the considered problem. The second goal was to provide an overview of the different scheduling problems on which the considered methods were applied. This makes it easy to obtain all the literature dealing with a specific problem variant, or to determine less investigated research areas.

Although the research in this area is already extensive, there are still several areas in which it can be improved. Additionally, recent years also saw the rise of some new problem variants that could become more important in the following years. Therefore, this study also outlines the limitations and possible future directions that can be performed in this area.

\ifCLASSOPTIONcaptionsoff
  \newpage
\fi



%

\bibliographystyle{IEEEtran}
\bibliography{IEEEexample}

%

\begin{IEEEbiography}[{\includegraphics[width=1in,height=1.25in,clip,keepaspectratio]{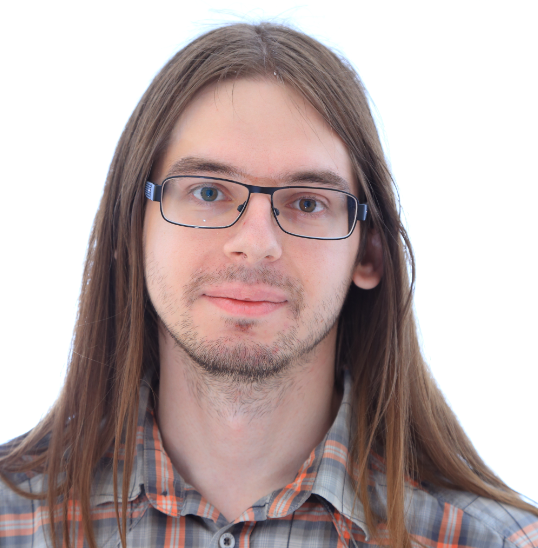}}]{Marko Đurasević}
received his Ph.D. degree in computer science from the University of Zagreb, Faculty of Electrical Engineering and Computing. He is currently an Assistant Professor at the same university. His main research interest is the application of hyper-heuristic methods for various scheduling problems. 
\end{IEEEbiography}

\begin{IEEEbiography}[{\includegraphics[width=1in,height=1.25in,clip,keepaspectratio]{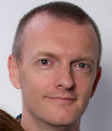}}]{Domagoj Jakobović}
is Full professor at Faculty of Electrical Engineering and Computing, University of Zagreb. He
received B.S. degree in December 1996 and MS degree
in December 2001 in Electrical Engineering. Since
1999 he is a member of the research and teaching staff
at the Department of Electronics, Microelectronics, Computer and Intelligent
Systems of Faculty of Electrical Engineering and Computing, 
University of Zagreb. He received
Ph.D. degree in December 2005 on the subject of generating scheduling heuristics with genetic programming.
\end{IEEEbiography}




\end{document}